\documentclass[runningheads]{llncs}

\usepackage[mobile]{eccv}

\usepackage{eccvabbrv}

\usepackage{graphicx}
\usepackage{booktabs}
\usepackage{arydshln}
\usepackage{colortbl}
\definecolor{graytext}{gray}{0.3}
\usepackage[accsupp]{axessibility}  

\usepackage{orcidlink}
\usepackage{multirow}
\usepackage{xspace}
\usepackage{xcolor}
\usepackage{amssymb}  
\usepackage{pifont}
\usepackage{cuted}
\usepackage{caption} 
\usepackage{wrapfig}
\usepackage{colortbl}
\newcommand{\cmark}{\textcolor{green!60!black}{\ding{51}}}  
\newcommand{\xmark}{\textcolor{red}{\ding{55}}}  

\usepackage{soul}

\newcommand\POEMs{POEM$_{\text{v1}}$\xspace}
\newcommand\POEMg{POEM$_{\text{v2}}$\xspace}

\usepackage{hyperref}
\hypersetup{
    colorlinks=true,
    urlcolor=eccvblue,  
}

\begin{document}


\title{HGGT: Robust and Flexible 3D Hand Mesh Reconstruction from Uncalibrated Images}
\titlerunning{HGGT}

\author{
Yumeng Liu\inst{1}
\and Xiao-Xiao Long\inst{2} \and 
 Marc Habermann\inst{3} \and 
 Xuanze Yang\inst{1} \\
 Cheng Lin\inst{4} \and 
 Yuan Liu\inst{5}\and 
 Yuexin Ma\inst{6}\and 
 Ligang Liu\inst{1}\textsuperscript{*}
}
\authorrunning{Y.~Liu et al.}

{
\renewcommand{\thefootnote}{}
\footnotetext{* Corresponding author.}
}

\institute{
University of Science and Technology of China \and 
Nanjing University \and 
Max-Planck-Institut für Informatik \and
Macau University of Science and Technology \and
Hong Kong University of Science and Technology \and
ShanghaiTech University
}

\maketitle
\begin{center}
    \vspace{-7mm}
    {\small\textcolor{eccvblue}{\textbf{Project Page:}} 
     \href{https://lym29.github.io/HGGT/}{\texttt{https://lym29.github.io/HGGT/}}}
    \vspace{1mm}
\end{center}

\begin{center}
    \centering
    \vspace{-10pt}
    \includegraphics[width=\textwidth]{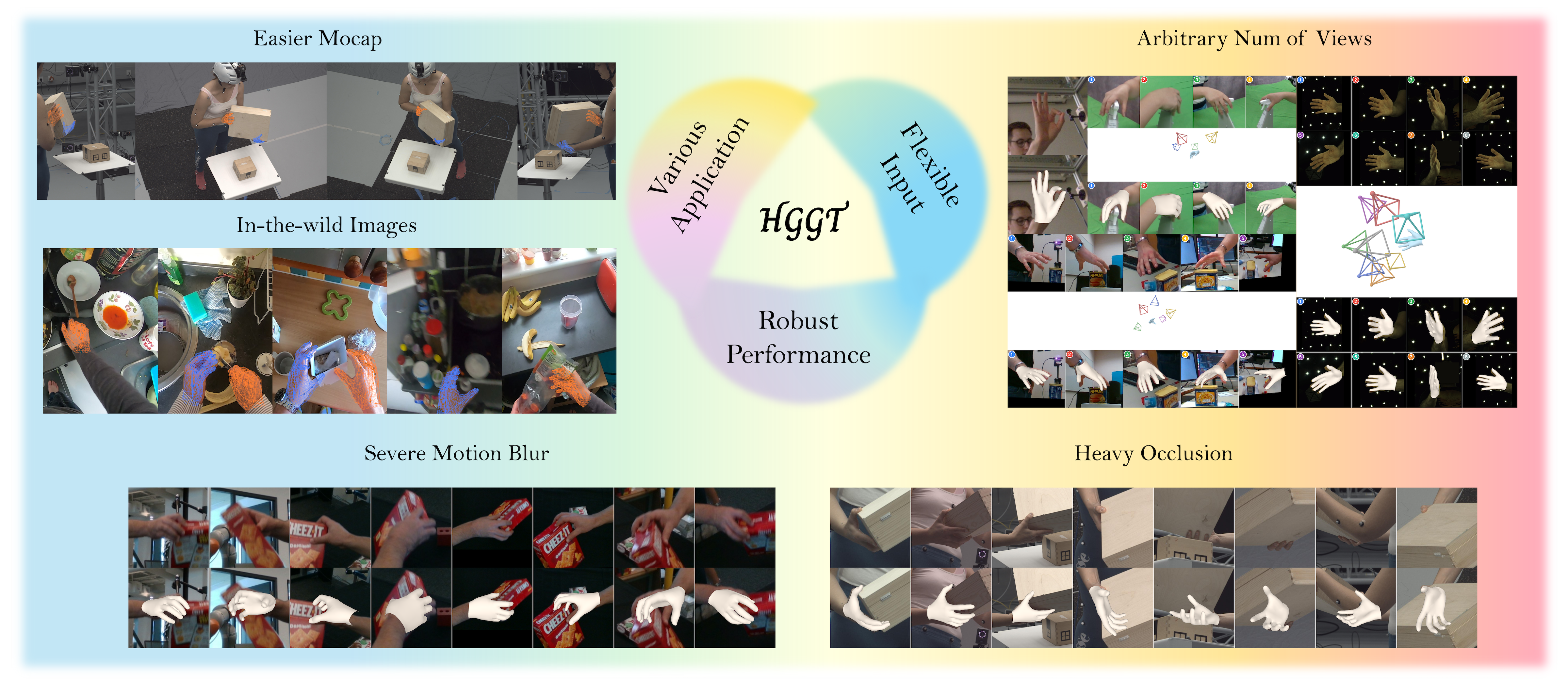} 
    \captionsetup{hypcap=false} 
    \captionof{figure}{ We introduce {\bf H}and {\bf G}eometry {\bf G}rounding {\bf T}ransformer (\textbf{HGGT}),  a scalable and generalized solution for 3D hand mesh recovery. Our method unifies diverse data sources to achieve robust performance across varying camera viewpoints and environments. 
    }
    \vspace{-1em}
    \label{fig:teaser}
\end{center}

\vspace{-1em}

\begin{abstract}
Recovering high-fidelity 3D hand geometry from images is a critical task in computer vision, holding significant value for domains such as robotics, animation and VR/AR. Crucially, scalable applications demand both accuracy and deployment flexibility, requiring the ability to leverage massive amounts of unstructured image data from the internet or enable deployment on consumer-grade RGB cameras without complex calibration. However, current methods face a dilemma. While single-view approaches are easy to deploy, they suffer from depth ambiguity and occlusion. Conversely, multi-view systems resolve these uncertainties but typically demand fixed, calibrated setups, limiting their real-world utility. To bridge this gap, we draw inspiration from 3D foundation models that learn explicit geometry directly from visual data. By reformulating hand reconstruction from arbitrary views as a visual-geometry grounded task, we propose a feed-forward architecture that, for the first time in literature, jointly infers 3D hand meshes and camera poses from uncalibrated views. Extensive evaluations show that our approach outperforms state-of-the-art benchmarks and demonstrates strong generalization to uncalibrated, in-the-wild scenarios. 
\end{abstract}
\section{Introduction}
\label{sec:intro}

The hand serves as our primary interface for interacting with the world. The detection and reconstruction of the hand has, thus, long been a pivotal task in computer vision, as it is key to a wide range of applications, such as analyzing human behavior~\cite{qi2024computer}, enhancing VR realism~\cite{pei2022hand,grauman2022ego4d}, and empowering robots with human-level dexterity~\cite{wang2025dexh2r}.

The advent of Vision Transformers (ViT) has recently revolutionized 3D human recovery. HaMeR\cite{pavlakos2024reconstructing} demonstrated that scaling up ViT backbones can significantly outperform traditional CNN-based regressors for monocular hand reconstruction. However, as a single-view method, it remains vulnerable to depth ambiguity and occlusion. To overcome these monocular limitations, POEM \cite{yang2024POEM} introduced a Point-Embedded Transformer that leverages multi-view geometry to fuse features onto a set of 3D basis points. However, POEM requires accurate camera calibration to perform feature aggregation, which limits its flexibility in in-the-wild, uncalibrated environments. 

In parallel, the broader 3D vision community has witnessed a paradigm shift towards feed-forward 3D reconstruction, exemplified by DUSt3R \cite{dust3r_cvpr24}, Mast3R \cite{mast3r_eccv24}, and VGGT \cite{wang2025vggt}. These foundation models demonstrate that scene attributes like camera extrinsics, depth, and point tracks can be learned directly from large-scale data via a unified transformer, obviating the need for traditional optimization. 
Inspired by this, concurrent works such as Human3R \cite{chen2025human3r} and UniSH \cite{li2026unish} have attempted to leverage these priors for human pose estimation. However, they typically treat the foundation model as a frozen module and focus solely on aligning human poses with the reconstructed 3D scene. 
Nevertheless, directly applying such scene-level models to human reconstruction or hand pose estimation in a feedforward manners remains relatively unexplored despite its potential benefits. We found that directly applying these methods to multi-view hand images performs poorly since these images exhibit minimal visual overlap between views. Therefore, these models often struggle to infer accurate camera poses, necessitating a more integrated approach that jointly reasons about hand geometry and camera pose.

To bridge this gap, we introduce a feedforward framework that empowers the VGGT backbone to tackle the challenge of 3D hand reconstruction from uncalibrated multi-view images. Specifically, we append a transformer decoder where learnable hand and camera tokens act as task-specific queries to retrieve geometric and view-dependent cues from the holistic image context via cross-attention. These decoded features are then routed into specialized heads to predict hand pose and camera parameters, respectively. This design effectively disentangles the representation of hand geometry from camera viewpoints, ensuring robust estimation.

To maximize the performance of our proposed framework, extensive data diversity is essential. 
Specifically, the model must learn to handle arbitrary camera setups and complex backgrounds, which requires training on a rich mix of scenarios. 
We achieve this by integrating three distinct types of data: 
First, we incorporate massive \textbf{in-the-wild monocular images} to expose the model to diverse environments and lighting conditions, thereby significantly enhancing its robustness and generalization in unconstrained scenarios.
Second, \textbf{real-world multi-view datasets} provide high quality 3D annotation, ensuring accurate supervision for geometric alignment.
Finally, to break free from the fixed camera rigs typical for studio collections, we introduce a \textbf{large-scale synthetic dataset} with randomized viewpoints; this is critical for driving the model to learn intrinsic geometric invariance rather than memorizing specific capture setups.

Extensive experiments on public benchmarks demonstrate the effectiveness of our proposed framework. Our model outperforms state-of-the-art methods in both 3D mesh recovery and 2D keypoint detection. 

In summary, our main contributions are as follows:
\begin{itemize}
    \item We propose the \textbf{first feed-forward framework} for calibration-free multi-view hand geometry recovery. Our unified transformer architecture jointly estimates camera poses and hand meshes, eliminating the need for pre-calibration or iterative optimization.
    
    \item We introduce a new \textbf{synthetic dataset} and a \textbf{mixed-data training strategy} that effectively leverages real monocular data, real multi-view data and synthetic multi-view data. The diversity of data resource significantly enhances the model's generalization capabilities across different domains.
    
    \item We conduct \textbf{extensive benchmarking} on open datasets, demonstrating that our method exhibits superior robustness in uncalibrated settings, outperforming existing solutions.
\end{itemize}

\section{Related Work}

\paragraph{Multi-view 3D Reconstruction.}
Reconstructing 3D geometry from multiple images, commonly known as multi-view stereo (MVS), is a fundamental problem in computer vision. MVS techniques aim to recover detailed 3D shapes by leveraging consistent features and geometric cues observed from multiple views.
Early traditional methods rely on handcrafted photometric consistency and propagation strategies~\cite{galliani2015massively, schonberger2016pixelwise}. 
The advent of deep learning introduced learning-based MVS, which integrates differentiable geometric modules like voxel-based cost volumes~\cite{yao2018mvsnet, gu2020cascade} or point-based aggregations~\cite{chen2019PointMVSNet}. 
These methods typically utilize camera parameters to perform explicit homography warping for feature alignment across views. 
Recently, a paradigm shift has occurred, pioneered by DUSt3R~\cite{dust3r_cvpr24} and MASt3R~\cite{mast3r_eccv24}, which directly estimate aligned dense point clouds from a pair of views. A series of follow-up works~\cite{cut3r, tang2024mv, zhang2025flare, yang2025fast3r} have further exemplified this paradigm.
VGGT~\cite{wang2025vggt} advances this by introducing a scalable framework capable of processing an arbitrary number of views jointly. 
It employs a novel Alternating Attention mechanism to capture dense multi-view correlations and directly infer camera parameters and 3D geometry in a single feed-forward pass. 
Leveraging this robust global geometric understanding, we adopt the VGGT backbone to drive our hand reconstruction. However, naively applying VGGT to uncalibrated in-the-wild images results in erroneous camera estimation and distorted point clouds, we will give a more detailed discussion in Section~\ref{sec:experiments}.

\paragraph{Hand Reconstruction.}
Recovering 3D hand pose and shape from images remains a fundamental challenge in computer vision. 
Recovering 3D hand pose and shape from images remains a fundamental challenge in computer vision. 
A significant body of work~\cite{panteleris2018using,zhang2019end,xiang2019monocular,zhou2020monocular} focuses on the monocular setting, \cite{hasson2019learning, pavlakos2024reconstructing, kong2022identity,potamias2025wilor} directly regressing the parameters of the differentiable MANO model~\cite{romero2022embodied} to constrain the solution space. Another line is non-parametric approach that directly regresses the vertices of the MANO mesh~\cite{choi2020pose2mesh,ge20193d,kulon2020weakly}.
To effectively extract features for accurate estimation from a single image, researchers have explored various architectures, ranging from early Graph Convolutional Networks (GCNs)~\cite{ge20193d,chen2021camera} to alternative representations such as voxels~\cite{moon2020i2l} or UV maps~\cite{chen2021i2uv}. 
Despite these architectural advancements, monocular methods remain fundamentally limited by scale-depth ambiguity. 
To address this, POEM~\cite{yang2024POEM} extended the Transformer paradigm to multi-view settings, modeling the hand mesh as a point set within a multi-view stereo framework. 
However, this approach heavily relies on pre-calibrated camera parameters to fuse visual features across different views.
In contrast, we tackle the more challenging uncalibrated setting. 
We demonstrate the effectiveness of a VGGT-based architecture in this context. 
By directly regressing MANO parameters and 2D keypoints, our approach validates that such a concise, direct paradigm yields robust performance, effectively resolving hand reconstruction problem without relying on complex intermediate representations or explicit camera calibration.

\section{Method}
\begin{figure}[t]
    \centering
    \includegraphics[width=\linewidth]{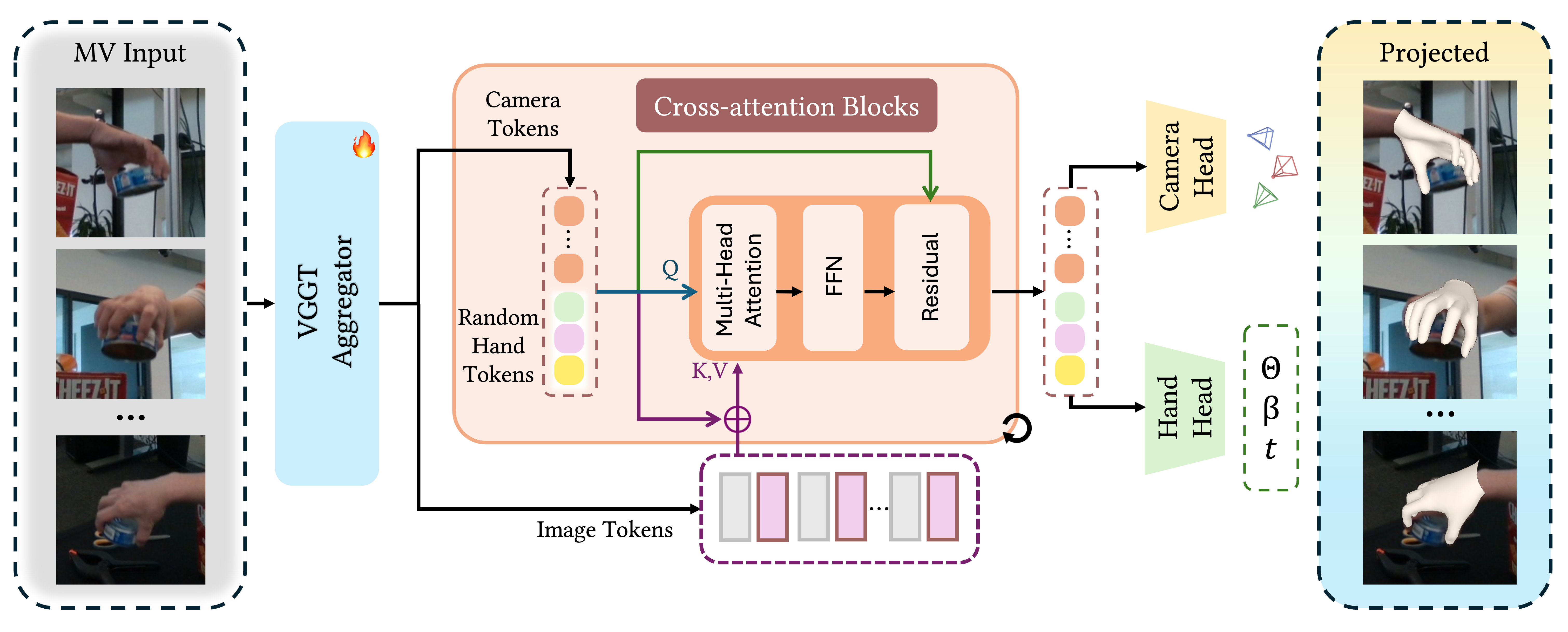}
    \caption{\textbf{The pipeline of HGGT.} Given uncalibrated multi-view images, we first employ a VGGT Aggregator to extract image tokens and initial camera tokens. These are processed alongside learnable random hand tokens via a series of Cross-attention Blocks. Finally, two parallel heads predict the camera parameters and the canonical MANO parameters ($\boldsymbol{\theta}, \boldsymbol{\beta}, \mathbf{t}$), which can be re-projected onto the input views for verification.}
    \vspace{-10pt}
    \label{fig:pipeline}
    \vspace{-10pt}
\end{figure}

We present \textbf{HGGT}, a feed-forward framework designed to estimate 3D hand pose and shape from uncalibrated multi-view images.
We begin by formalizing the problem setting in Section~\ref{sec:problem_definition}.
Following this, we describe the data preparation process in Section~\ref{sec:dataset_preparation}.
The core architecture and design choices of our model are detailed in Section~\ref{sec:model_design} with an illustration of our pipeline in~\autoref{fig:pipeline}.
Finally, we discuss the optimization objectives in Section~\ref{sec:loss_function} and the training strategy in Section~\ref{sec:training_strategy}.

\subsection{Problem Definition}
\label{sec:problem_definition}
Our goal is to estimate the 3D hand mesh and camera parameters from an arbitrary number of RGB images without prior calibration. Formally, given a set of $S$ images $\{\mathbf{I}_s\}_{s=1}^{S}$ captured from unknown viewpoints, HGGT jointly predicts: \textit{(i)} the MANO hand parameters $(\boldsymbol{\theta}, \boldsymbol{\beta}, \mathbf{t})$ defined in the first camera coordinate system; and \textit{(ii)} view-specific camera pose encodings $\{\mathbf{p}_s\}_{s=1}^{S}$. Here, $\boldsymbol{\theta} \in \mathbb{R}^{48}$ represents the articulated hand pose in axis-angle format, where the first three dimensions correspond to the global orientation; $\boldsymbol{\beta} \in \mathbb{R}^{10}$ denotes the shape parameters; and $\mathbf{t} \in \mathbb{R}^{3}$ represents the global translation. To this end, we employ distinct prediction heads atop a shared feature backbone, enabling effective multi-task learning and facilitating mutual supervision across tasks.

\subsection{Dataset Preparation}
\label{sec:dataset_preparation}

We construct a hybrid training dataset by amalgamating three distinct data sources to ensure coverage of both semantic diversity and geometric precision. 
First, we incorporate large-scale in-the-wild monocular image collections~\cite{pavlakos2024reconstructing} to encompass a broad spectrum of illumination conditions and background complexities. 
Second, we utilize established real-world multi-view datasets~\cite{yang2024POEM} for their high-fidelity 3D annotations; however, acknowledging that these are predominantly captured via fixed rigs with limited extrinsic variation, we further augment our training set with a synthetic component. 

\begin{figure}[htp]
    \centering
    \includegraphics[width=\linewidth]{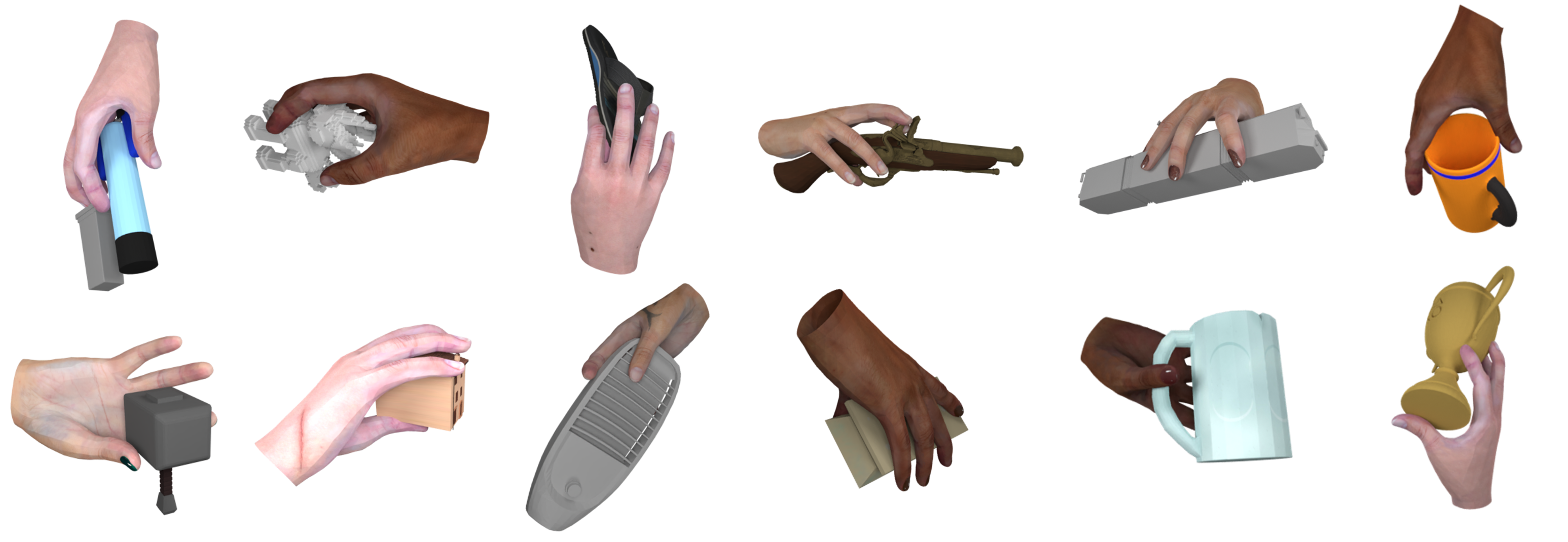}
    \caption{
    \textbf{Samples from our synthetic dataset.} 
    It contains diverse photorealistic hand-object interactions. 
    }
    \label{fig:dataset_samples}
    \vspace{-20pt}
\end{figure}

Specifically, to introduce randomized camera settings absent in real-world captures, we leverage the pre-computed grasp interactions from GraspXL~\cite{zhang2024graspxl} (based on Objaverse objects~\cite{deitke2023objaverse}). We subsequently render these interactions with photorealistic skin textures using the asserts provided by Dart~\cite{gao2022dart} to minimize the synthetic-to-real domain gap, as visualized in Fig.~\ref{fig:dataset_samples}. 
This synthetic subset significantly expands our data scale, comprising 85,683 image sets for training and 21,242 for validation, with each set containing 10 rendered views.

\subsection{Model Design}
\label{sec:model_design}
An illustration of our pipeline is shown in~\autoref{fig:pipeline}. 
The pipeline originates with the VGGT aggregator, which encodes multi-view inputs to simultaneously extract image features and initialize camera tokens. These representations are then fed into a \textbf{Unified Cross-attention Refinement Module} alongside a set of learnable random hand tokens, where both the camera and hand tokens are iteratively refined.
This mechanism allows the model to effectively aggregate visual cues of hand across different views.
Finally, the updated tokens are decoded by task-specific heads: a \textbf{Camera Head} that predicts the extrinsic parameters for each view, and a \textbf{Hand Head} that regresses the MANO parameters $(\boldsymbol{\theta}, \boldsymbol{\beta}, \mathbf{t})$ to recover the 3D hand geometry.

\subsubsection{Multi-View Feature Aggregation.}
Our backbone utilizes the VGGT~\cite{wang2025vggt} alternating-attention architecture to extract multi-view features. It processes image patches and a learnable \emph{camera token} through interleaved frame-level and global attention layers. We resume the pretrained weight from VGGT and unfreeze all the layer of its aggregator to train on our task.

\subsubsection{Unified Cross-Attention Refinement.}
Following the feature extraction stage, we introduce a set of learnable latent embeddings, referred to as \textbf{hand tokens}, to represent the global hand geometry. These tokens are concatenated with the camera tokens (initialized by the VGGT Aggregator) and processed by 4 stacked cross-attention blocks.
In this mechanism, the task-specific tokens (i.e., both hand and camera tokens) function as Queries ($\mathbf{Q}$), while the encoded multi-view image features serve as Keys ($\mathbf{K}$) and Values ($\mathbf{V}$).
This design allows the hand tokens to explicitly query and aggregate relevant geometric cues from all uncalibrated views, iteratively refining their representations alongside the camera tokens using cross-view information.

\subsubsection{Parametric Hand Head.}
\label{sec:mano_head}
To recover the dense 3D mesh, we decode the refined hand tokens into the parameters of the MANO model~\cite{romero2022embodied}.
Specifically, the updated hand tokens are fed into a lightweight Multi-Layer Perceptron (MLP) module named \textbf{Hand Head}, which directly regresses the pose $\boldsymbol{\theta}$, shape $\boldsymbol{\beta}$, and translation $\mathbf{t}$ parameters.
These predicted parameters are subsequently passed through the differentiable MANO layer to generate the final 3D hand vertices $\mathcal{V} \in \mathbb{R}^{778 \times 3}$ and joints $\mathcal{J} \in \mathbb{R}^{21 \times 3}$.

\subsubsection{Camera Pose Estimation Head}
\label{sec:cam_head}
The camera head estimates the extrinsic and intrinsic parameters for each view relative to the first frame. Specifically, it predicts a 9-dimensional encoding $\mathbf{p}_s = [\mathbf{T}_s, \mathbf{q}_s, \mathbf{f}_s] \in \mathbb{R}^9$ for each view $s$, comprising the translation $\mathbf{T}_s \in \mathbb{R}^3$, a unit quaternion rotation $\mathbf{q}_s \in \mathbb{R}^4$, and Field of View (FoV) $\mathbf{f}_s \in \mathbb{R}^2$.
This head shares the same architecture as the camera head in VGGT~\cite{wang2025vggt}.

\subsection{Loss Function}
\label{sec:loss_function}
Our training objective is designed to decouple the optimization of intrinsic hand properties from the geometric imaging process. The total loss $\mathcal{L}_{\text{total}}$ is composed of three distinct components:
\paragraph{Hand Loss.}
This component focuses exclusively on the reconstruction of the hand mesh in canonical 3D space, independent of the camera viewpoint. We supervise the MANO parameters ($\boldsymbol{\theta}, \boldsymbol{\beta}$) and the resulting root-relative 3D joint coordinates $J_{3D}$. However, since all of single-view datasets lack ground-truth MANO annotations and some of them may only contain 2D annotation lacking 3D joints supervision~\cite{pavlakos2024reconstructing}, we apply supervision only when available. Consequently, the loss is formulated as:
\begin{equation}
    \mathcal{L}_{\text{hand}} = \mathcal{I}_{\text{mano}} \left( \|\boldsymbol{\theta} - \hat{\boldsymbol{\theta}}\|_2^2 + \|\boldsymbol{\beta} - \hat{\boldsymbol{\beta}}\|_2^2 \right) + \mathcal{I}_{\text{joints}}\|J_{3D} - \hat{J}_{3D}\|_2^2,
\end{equation}
where $\mathcal{I}_{\text{mano}}$ and $\mathcal{I}_{\text{joints}}$ are an indicator function that equals 1 if ground-truth MANO parameters or 3D joints are provided, and 0 otherwise (e.g., for single-view samples).

\paragraph{Camera Loss.}
To ensure geometric consistency between the reconstructed 3D hand and the 2D image plane, we introduce a camera alignment loss. Since relative camera pose estimation relies on multi-view cues, this loss is activated exclusively for samples with multiple views ($S > 1$). Specifically, we enforce explicit constraints on the camera extrinsics as follows:
\begin{equation}
    \mathcal{L}_{\text{cam}} = \dfrac{1}{S}\sum_{s=1}^S (\|\mathbf{T}_s - \hat{\mathbf{T}}_s\|_2^2 + \|R(\mathbf{q}_s) \ominus R(\hat{\mathbf{q}_s})\|_2^2+||\mathbf{f}_s - \hat{\mathbf{f}}_s||_2^2),
\end{equation}

where $\hat{\cdot}$ denotes the predicted parameters, $\mathcal{R}(\cdot)$ is the rotation matrix derived from the quaternion, and $\ominus$ denotes the geodesic distance on $SO(3)$.

\paragraph{Projection Consistency Loss.} 
To bridge the gap between the predicted 3D structure and the observed 2D images, we employ a projection consistency loss. This is particularly crucial for single-view samples where ground-truth 3D annotations are unavailable; in such cases, 2D reprojection serves as the sole supervision to align the predicted hand and camera. The loss comprises two key components:

\begin{enumerate}
    \item \textbf{Reprojection Alignment:} We project the \textit{predicted} 3D joints onto the image plane using the \textit{predicted} camera intrinsics and extrinsics. By minimizing the distance between these projected points and the ground-truth 2D keypoints, we simultaneously constrain the 3D hand geometry and the camera pose:
    \begin{equation}
        \mathcal{L}_{\text{reproj}} = \frac{1}{S \cdot J} \sum_{i=1}^{S} \| \Pi(J_{3D}^{\text{pred}}, K_{\text{pred}}, [R|T]_{\text{pred}}) - J_{2D}^{\text{GT}} \|_2^2.
    \end{equation}
    This term ensures that the reconstructed 3D hand, when viewed through the predicted camera, visually matches the evidence in the input image.

    \item \textbf{Negative Depth Penalty:} To prevent degenerate solutions where points are mathematically projected but physically located behind the camera, we impose a negative depth penalty. Given the predicted depth $z_{i,j}$ for the $j$-th keypoint in the $i$-th view, this penalty is defined as:
    \begin{equation}
        \mathcal{L}_{\text{neg}} = \frac{1}{S \cdot J} \sum_{i=1}^{S} \sum_{j=1}^{J} \left( \max(0, -z_{i,j}) \right)^2,
    \end{equation}
    where $S$ is the number of views and $J$ is the number of hand keypoints.
\end{enumerate}

Finally, the total loss function is defined as:
\begin{equation}
\mathcal{L}_{\text{total}} = \sum_{l=1}^{L} \gamma^{L-l} ( \lambda_{\text{hand}} \mathcal{L}_{\text{hand}}^{(l)} + \lambda_{\text{cam}} \mathcal{L}_{\text{cam}}^{(l)} ) + \lambda_{\text{proj}} \mathcal{L}{\text{proj}}^{(L)},
\end{equation}
where $L$ denotes the total number of cross-attention blocks, and the superscript $(l)$ indicates the predictions from the $l$-th block. We apply intermediate supervision to the hand and camera parameters with exponentially increasing stage weights controlled by $\gamma = 0.6$. The 2D projection loss $\mathcal{L}_{\text{proj}}$, however, is applied exclusively to the final output to ensure terminal geometric consistency.

\subsection{Training Strategy}
\label{sec:training_strategy}
A major challenge in 3D hand reconstruction is the scarcity of large-scale datasets with diverse viewpoints and 3D annotations. While multi-view datasets provide essential geometric cues, they are often limited in scale. Conversely, single-view datasets are abundant but typically contain only 2D keypoint annotations. To maximize our model's performance and generalization capability, it is crucial to harness the complementary strengths of these data sources.

However, integrating these datasets in a unified training framework is non-trivial due to structural heterogeneity. 
First, the \textbf{variable input cardinality} prevents standard uniform batching, where single-view samples ($S=1$) coexist with multi-view samples ($S \in [2, 24]$). 
Second, this leads to \textbf{computational imbalance}, as resource consumption scales linearly with view counts, rendering static batching strategies inefficient. 
Finally, the \textbf{training objectives differ}: single-view data assumes a fixed canonical camera, whereas multi-view data necessitates additional camera and projection losses to enforce geometric consistency.

To address these issues, we employ a specialized training strategy via gradient accumulation. By accumulating gradients over alternating single view and multi-view steps, we simulate a joint batch that balances these heterogeneous objectives within a single weight update. Following the efficient protocol of VGGT~\cite{wang2025vggt}, we enforce a constant image budget $N_{\text{img}}$ to maximize hardware utilization, dynamically adjusting the batch size to $B = \lfloor N_{\text{img}} / S \rfloor$ for each step. By accumulating gradients over alternating single view and multi-view steps, we simulate a joint batch that optimizes both objectives simultaneously.

\section{Experiments}
\label{sec:experiments}
\subsection{Implementation Details}
We implement our framework using PyTorch, training a model with approximately 1.4 billion parameters on a cluster of 8 NVIDIA A100 GPUs. 
The optimization utilizes AdamW~\cite{loshchilov2017decoupled} with a weight decay of 0.05. 
To ensure stable convergence given the memory constraints, we set the micro-batch size to 32 images per GPU and perform gradient accumulation over 4 steps before each weight update. 
The learning rate follows a composite schedule: it linearly warms up from $1 \times 10^{-8}$ to $1 \times 10^{-6}$ during the first 5\% of the 80,000 total iterations, followed by a cosine decay to $1 \times 10^{-8}$, with the entire training process taking approximately 5 days. 
We employ an alternating training strategy between single-view and multi-view data. For single-view data, we fix the number of views to $N = 1$. For multi-view data, the maximum number of sampled frames depends on the available camera views provided by each dataset. Specifically, we dynamically sample $N \in [2, 10]$ frames per sequence for our synthetic dataset, $N \in [2, 5]$ for HO3D, $N \in [2, 8]$ for DexYCB, ARCTIC, and InterHand, and $N \in [2, 4]$ for OakInk.
Input images are processed with a patch size of 14 and resized to a maximum dimension of 518 pixels, with aspect ratios randomized between 0.5 and 1.0. 
Finally, to enhance robustness, we apply photometric augmentations same as VGGT~\cite{wang2025vggt}, including color jittering, Gaussian blur, and grayscale conversion.
During evaluation, we use the original image order to evaluate our method.

\subsection{Training Data Details}
\label{sec:training_data}

To ensure robust generalization and accurate 3D hand reconstruction across diverse scenarios, we train our model on a comprehensive mixture of single-view and multi-view datasets, encompassing both real-world captures and synthetic renderings.

\subsubsection{Single-View Datasets}
For single-view supervision, we adopt the extensive data collection curated by HaMeR~\cite{pavlakos2024reconstructing}. This compilation consolidates multiple datasets providing 2D or 3D hand annotations, specifically FreiHAND~\cite{zimmermann2019freihand}, HO3D~\cite{hampali2020honnotate}, MTC~\cite{xiang2019monocular}, RHD~\cite{zimmermann2017learning}, InterHand2.6M~\cite{Moon_2020_ECCV_InterHand26M}, H2O3D~\cite{hampali2020honnotate}, DEX YCB~\cite{chao2021dexycb}, COCO WholeBody~\cite{jin2020whole}, Halpe~\cite{fang2022alphapose}, and MPII NZSL~\cite{simon2017hand}. This yields approximately 2.7M training examples. While the majority of these images are captured in controlled environments, around $5\%$ of the training examples (from COCO WholeBody, Halpe, and MPII NZSL) consist of in-the-wild images, which significantly enhances the model's robustness to diverse backgrounds and lighting conditions.

\subsubsection{Multi-View Real Datasets}
For real-world multi-view data, we utilize the standardized collection introduced by POEM~\cite{yang2024POEM}. This integrates observations and 3D annotations from multiple calibrated cameras across several prominent hand-object interaction datasets, including DexYCB~\cite{chao2021dexycb}, HO3D~\cite{hampali2020honnotate}, OakInk~\cite{yang2022oakink}, ARCTIC~\cite{fan2023arctic}, and InterHand2.6M~\cite{Moon_2020_ECCV_InterHand26M}. In total, this compilation provides approximately 438K multi-view training frames (comprising roughly 25K, 9K, 58K, 135K, and 210K frames from the respective datasets). 

\subsubsection{Multi-View Synthetic Datasets}
A major limitation of existing real-world multi-view datasets is that they are typically captured under fixed camera configurations, leading to a severe lack of viewpoint diversity. To address this issue, we synthesize a large-scale multi-view dataset that provides highly diverse camera viewpoints. Specifically, we render hand-object interaction sequences utilizing the rich kinematic and geometric data from GraspXL~\cite{zhang2024graspxl}. To ensure high visual realism and appearance diversity, we apply a wide variety of high-quality object materials sourced from the Objaverse~\cite{deitke2023objaverse} and DART~\cite{gao2022dart} datasets during the rendering process. This synthetic subset significantly expands our data scale, comprising 85,683 multi-view frames for training and 21,242 for validation. With each frame containing 10 rendered views, this yields a substantial total of 856,830 synthetic training images. In conclusion, this synthetic data serves as a valuable complement to existing real-world datasets, providing diverse camera viewpoints, realistic rendering results, and rich annotations (including hand pose, as well as object and hand masks and depth maps).

\begin{figure}[htbp]
    \centering
    \begin{minipage}{0.45\linewidth}
        \centering
        \includegraphics[width=\linewidth]{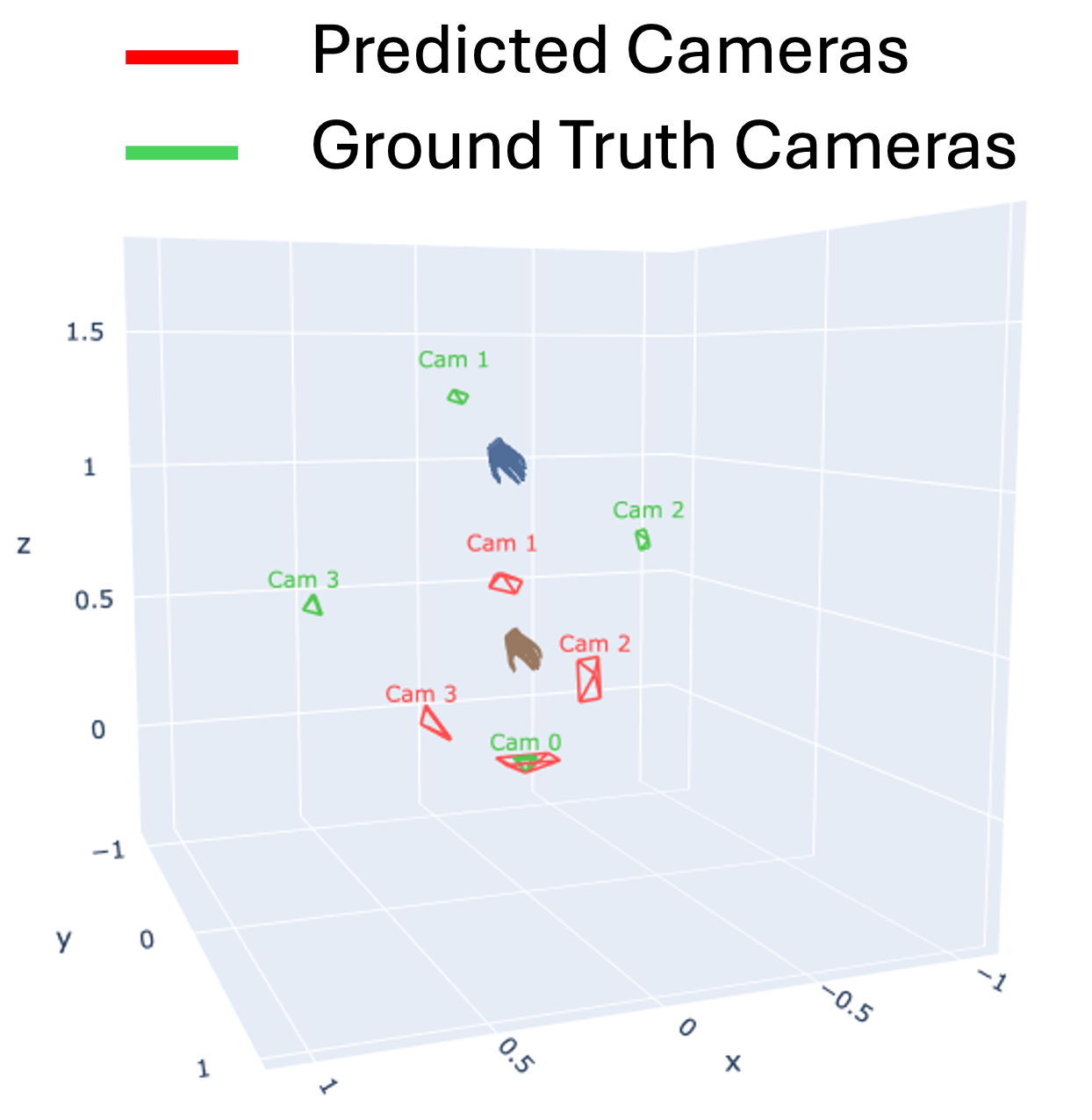}
        \caption{Camera Layout Visualization.}
        \label{fig:camera_example}
    \end{minipage}
    \hfill 
    \begin{minipage}{0.5\linewidth}
        \centering
        \includegraphics[width=\linewidth]{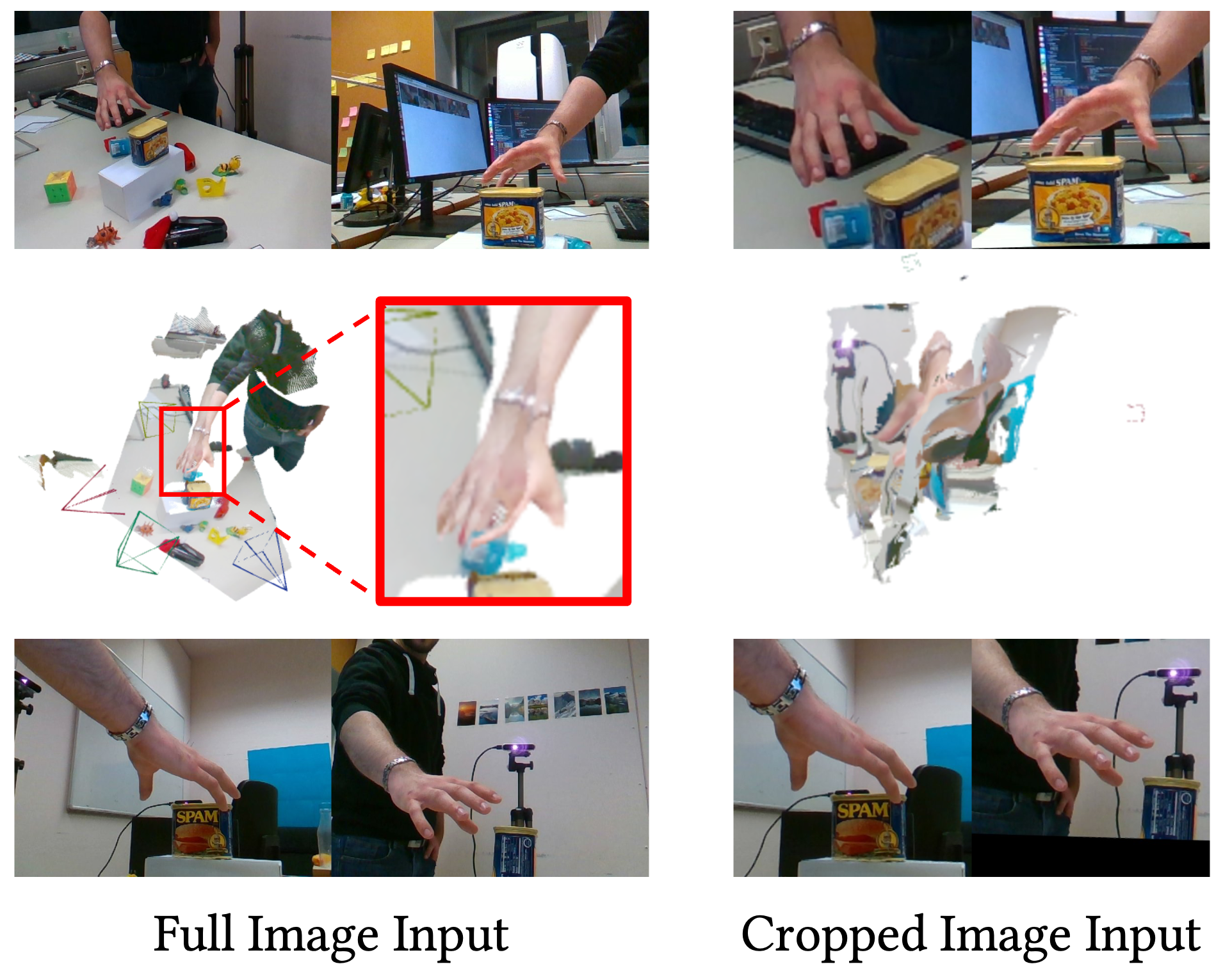}
        \caption{Failure cases of off-the-shelf VGGT on HO3D datasets.} 
        \label{fig:vggt_failure}
    \end{minipage}
    \vspace{-20pt}
\end{figure}

\subsection{Evaluation Metric}
We report the Mean Per Joint Position Error (\textbf{MPJPE}) and Mean Per Vertex Position Error (\textbf{MPVPE}) in millimeters. 
In a our setting, the absolute depth of the wrist is inherently unobservable due to scale ambiguity. 
While our predicted camera layout is structurally consistent with the ground truth, as shown in the example of~\autoref{fig:camera_example}, aligning the global coordinate system requires known wrist depth, which is mathematically equivalent to the Root-Relative (RR) evaluation setting. 
Therefore, to ensure a fair and valid comparison, we strictly report results under root-relative alignment and Procrustes analysis (PA). 
Additionally, we evaluate the Area Under the Curve (AUC) of the Percentage of Correct Keypoints (PCK) to assess prediction robustness.

\subsection{Limitations of Off-the-shelf VGGT.} 
We observe that directly feeding multi-view hand images into the off-the-shelf VGGT results in severe misalignment, regardless of the input strategy. As shown in~\autoref{fig:vggt_failure}, when using cropped images (the standard protocol for hand reconstruction), the minimal visual overlap between sparse views causes the geometric estimation to collapse entirely. 
However, simply reverting to full-frame images does not resolve the issue. 
As shown in the left of~\autoref{fig:vggt_failure}, although the background context helps establish a rough global coordinate system, the specific point cloud of the hand remains poorly aligned. Our quantitative analysis in~\autoref{tab:main_table} also confirms that the direct use of off-the-shelf VGGT leads to sub-optimal results for current state-of-the-art methods.
This is because VGGT's camera estimation is dominated by static, high-texture background features, failing to capture the fine-grained geometric consistency required for the relatively small and weakly textured hand region.

\begin{table*}[t]
\begin{center}
\caption{The ``Cali-Free'' column denotes calibration dependency. Rows in \textcolor{graytext}{gray} (marked with \textcolor{red}{\xmark}) represent methods utilizing ground-truth camera parameters, serving as an \textit{upper-bound reference}. Rows in black (marked with \textcolor{green}{\cmark}) represent calibration-free approaches. Within the calibration-free setting, we highlight the \textbf{best} results in bold and the \underline{second-best} with an underline.}
\vspace{-10pt}

\label{tab:main_table}
\footnotesize

\renewcommand{\arraystretch}{1.15}
\setlength{\dashlinedash}{0.6pt}
\setlength{\dashlinegap}{1.8pt}
\addtolength{\tabcolsep}{-1pt}

\resizebox{\linewidth}{!}{
\begin{tabular}{llcccccccccc}
\toprule

\multirow{2}{*}{datasets} & \multirow{2}{*}{\#} & \multirow{2}{*}{methods} & \multirow{2}{*}{Cali-Free} 
& \multicolumn{4}{c}{Root Relative} 
& \multicolumn{4}{c}{Procrustes Alignment} \\

\cmidrule(lr){5-8}
\cmidrule(lr){9-12}

& & & 
& MPVPE$_{\downarrow}$ & AUC$_{V20\uparrow}$ & MPJPE$_{\downarrow}$ & AUC$_{J20\uparrow}$
& MPVPE$_{\downarrow}$ & AUC$_{V20\uparrow}$ & MPJPE$_{\downarrow}$ & AUC$_{J20\uparrow}$ \\
\midrule

\multirow{5}*{\shortstack[l]{DexYCB-Mv \\ (8 views)}} 
& \texttt{1}  & \POEMs & \xmark & \color{graytext}7.21 & \color{graytext}0.65 & \color{graytext}7.30 & \color{graytext}0.62 & \color{graytext}4.00 & \color{graytext}0.80 & \color{graytext}3.93 & \color{graytext}0.78 \\
& \texttt{2}  & \POEMg-large & \xmark & \color{graytext}7.50 & \color{graytext}0.64 & \color{graytext}7.54 & \color{graytext}0.63 & \color{graytext}3.84 & \color{graytext}0.81 & \color{graytext}3.68 & \color{graytext}0.81 \\
\cdashline{3-12}
& \texttt{3}  & VGGT+POEM & \cmark  & 39.08 & 0.10 & 39.99 & 0.10 & 11.06 & 0.54 & 10.20 & 0.53 \\
& \texttt{4}  & Ours+POEM & \cmark  & \underline{19.69} & \underline{0.29} & \underline{20.18} & \underline{0.27} & \textbf{4.38} & \underline{0.78} & \textbf{4.19} & \textbf{0.79} \\
& \texttt{5}  & Ours & \cmark & \textbf{13.29} & \textbf{0.41} & \textbf{13.79} & \textbf{0.39} & \underline{4.47} & \textbf{0.78} & \underline{4.57} & \underline{0.77} \\
\midrule
\multirow{5}*{\shortstack[l]{OakInk-Mv \\ (4 views)}}   
& \texttt{6} & {\POEMs} & \xmark & \color{graytext}7.63 & \color{graytext}0.64 & \color{graytext}7.46 & \color{graytext}0.63 & \color{graytext}4.21 & \color{graytext}0.79 & \color{graytext}4.00 & \color{graytext}0.78 \\
& \texttt{7} & \POEMg-large & \xmark & \color{graytext}8.47 & \color{graytext}0.59 & \color{graytext}8.47 & \color{graytext}0.58 & \color{graytext}4.49 & \color{graytext}0.78 & \color{graytext}4.24 & \color{graytext}0.77 \\
\cdashline{3-12}
& \texttt{8}  & VGGT+POEM &  \cmark  & 47.29 & 0.12 & 48.66 & 0.12 & 13.18 & 0.48 & 12.08 & 0.47 \\
& \texttt{9}  & Ours+POEM & \cmark & \underline{26.58} & \underline{0.22} & \underline{27.34} & \underline{0.21} & \textbf{5.49} & \textbf{0.73} & \textbf{5.14} & \textbf{0.72} \\
& \texttt{10} & Ours & \cmark & \textbf{20.13} & \textbf{0.37} & \textbf{20.59} & \textbf{0.37} & \underline{6.14} & \underline{0.70} & \underline{6.12} & \underline{0.70} \\
\midrule

\multirow{4}*{\shortstack[l]{InterHand-Mv \\ (8 views)}} 
& \texttt{11} & \POEMg-large & \xmark & \color{graytext}9.04 & \color{graytext}0.62 & \color{graytext}8.82 & \color{graytext}0.60 & \color{graytext}5.23 & \color{graytext}0.76 & \color{graytext}4.72 & \color{graytext}0.70 \\
\cdashline{3-12}
& \texttt{12}  & VGGT+POEM & \cmark & 49.92 & 0.07 & 50.57 & 0.08 & 20.16 & 0.27 & 18.49 & 0.23 \\
& \texttt{13}  & Ours+POEM & \cmark & \textbf{19.18} & \textbf{0.29} & \textbf{19.47} & \textbf{0.27} & \textbf{5.98} & \textbf{0.73} & \textbf{5.37} & \textbf{0.69} \\
& \texttt{14} & Ours & \cmark & \underline{23.22} & \underline{0.20} & \underline{27.46} & \underline{0.20} & \underline{6.05} & \underline{0.71} & \underline{13.81} & \underline{0.59} \\
\midrule

\multirow{4}*{\shortstack[l]{ARCTIC-Mv \\ (8 views)}}    
& \texttt{15} & \POEMg-large & \xmark & \color{graytext}7.13 & \color{graytext}0.64 & \color{graytext}6.80 & \color{graytext}0.65 & \color{graytext}4.43 & \color{graytext}0.78 & \color{graytext}3.91 & \color{graytext}0.78 \\
\cdashline{3-12}
& \texttt{16}  & VGGT+POEM & \cmark       & 20.19 & \underline{0.25} & 20.67 & \underline{0.24} & 8.21 & 0.60 & 7.29 & 0.62 \\
& \texttt{17}  & Ours+POEM & \cmark & \underline{19.67} & \textbf{0.25} & \textbf{19.93} & \textbf{0.25} & \textbf{5.35} & \textbf{0.73} & \textbf{4.75} & \textbf{0.74} \\
& \texttt{18} & Ours & \cmark & \textbf{19.48} & 0.21 & \underline{20.04} & 0.19 & \underline{6.57} & \underline{0.67} & \underline{6.61} & \underline{0.67} \\
\midrule

& & & & & AUC$_{V50\uparrow}$ & & AUC$_{J50\uparrow}$ & & AUC$_{V50\uparrow}$ & & AUC$_{J50\uparrow}$ \\
\multirow{5}*{\shortstack[l]{HO3D-Mv \\ (5 views)}}     
& \texttt{19} & {\POEMs} & \xmark & \color{graytext}21.45 & \color{graytext}0.58 & \color{graytext}21.94 & \color{graytext}0.55 & \color{graytext}9.97 & \color{graytext}0.80 & \color{graytext}9.60 & \color{graytext}0.78 \\
& \texttt{20} & \POEMg-large& \xmark & \color{graytext}10.58 & \color{graytext}0.79 & \color{graytext}10.58 & \color{graytext}0.79 & \color{graytext}4.59 & \color{graytext}0.91 & \color{graytext}4.13 & \color{graytext}0.91 \\
\cdashline{3-12}
& \texttt{21}  & VGGT+POEM & \cmark & 56.07 & 0.23 & 57.53 & 0.20 & 21.05 & 0.59 & 20.29 & 0.47 \\
& \texttt{22} & Ours+POEM & \cmark & \underline{28.48} & \underline{0.48} & \underline{29.37} & \underline{0.46} & \underline{6.05} & \underline{0.88} & \underline{5.57} & \underline{0.88} \\
& \texttt{23} & Ours & \cmark & \textbf{9.98} & \textbf{0.80} & \textbf{10.03} & \textbf{0.80} & \textbf{3.30} & \textbf{0.93} & \textbf{3.33} & \textbf{0.93} \\
\bottomrule

\end{tabular}
}
\vspace{-15pt}
\end{center}
\end{table*}



\subsection{Main Results}
We present the quantitative comparison with the state-of-the-art method POEM in~\autoref{tab:main_table}. 
It is important to note that POEM relies on ground-truth (GT) camera parameters as input, while our method aims for reconstruction from uncalibrated images. 
To enable a fair evaluation under practical settings, we additionally evaluate POEM using predicted cameras from two sources: (1) the original VGGT estimation, and (2) our proposed camera prediction module.

\noindent\textbf{Comparison with GT-based Methods.} 
Despite operating without ground-truth camera information, our method achieves performance comparable to POEM (with GT inputs). 
Notably, on the HO3D dataset, our approach even surpasses the GT-aided POEM, demonstrating the robustness of our learned geometry. 

\noindent\textbf{Comparison under Uncalibrated Settings.} 
When POEM is restricted to using cameras predicted by off-the-shelf VGGT, our method significantly outperforms it. 
As shown in~\autoref{tab:main_table}, equipping POEM with our estimated cameras yields better results than the VGGT baseline. 
Despite this improvement, our method still surpasses POEM across the majority of benchmarks, with only a marginal deficit observed on the OakInk dataset. 
This highlights the robustness of our pose estimation framework, demonstrating its ability to maintain high accuracy across diverse geometric scenarios despite the challenges of uncalibrated inputs.

\begin{table}[htp] 
    \centering
    \caption{\textbf{Ablation Studies.} We analyze the impact of (a) freezing different stages of the VGGT backbone and (b) incorporating our rendered synthetic dataset. The default setting is highlighted in \colorbox{gray!20}{gray}.}
    \label{tab:ablation_combined}
    \resizebox{\linewidth}{!}{ 
        \begin{tabular}{l|c|c|cccc|cccc}
            \toprule
            \multirow{2}{*}{\textbf{Experiment}} & \multirow{2}{*}{\textbf{Setting}} & \multirow{2}{*}{\textbf{Trainable Params}} 
            & \multicolumn{4}{c|}{\textbf{Root Relative}} 
            & \multicolumn{4}{c}{\textbf{Procrustes Alignment}} \\
            \cmidrule(lr){4-7} \cmidrule(lr){8-11}
            & & & \textbf{MPVPE}$\downarrow$ & \textbf{AUC}$_{V50}\uparrow$ & \textbf{MPJPE}$\downarrow$ & \textbf{AUC}$_{J50}\uparrow$
            & \textbf{MPVPE}$\downarrow$ & \textbf{AUC}$_{V50}\uparrow$ & \textbf{MPJPE}$\downarrow$ & \textbf{AUC}$_{J50}\uparrow$ \\
            \midrule
            
            \multicolumn{11}{l}{\textit{(a) VGGT Freezing Strategy (on Real+Syn Data)}} \\
            \midrule
            \multirow{3}{*}{Freeze Layers} 
            & 24 & 0.46B &25.87  &0.51  &26.41  &0.50  &17.82  &0.65  &18.19  &0.64  \\
            & 20 & 0.56B &18.16  &0.65  &17.60  & 0.66 &13.72  &0.73  &14.06  &0.72  \\
            & 10 & 0.81B  &12.57  &0.75  &12.88  &0.74  &4.32  &0.92  &4.39  &0.92 \\
            & \cellcolor{gray!20}\textbf{None} & 1.37B & \cellcolor{gray!20}\textbf{9.98} & \cellcolor{gray!20}\textbf{0.80} & \cellcolor{gray!20}\textbf{10.03} & \cellcolor{gray!20}\textbf{0.80} & \cellcolor{gray!20}\textbf{3.30} & \cellcolor{gray!20}\textbf{0.93} & \cellcolor{gray!20}\textbf{3.33} & \cellcolor{gray!20}\textbf{0.93} \\
            \midrule
            
            \multicolumn{11}{l}{\textit{(b) Training Data Composition}} \\
            \midrule
            \multirow{2}{*}{Data Source} 
            & Real Only          & - & 15.30 &0.61  &15.33  &0.61  &5.97  &0.89  &6.04  &0.88  \\
            & \cellcolor{gray!20}\textbf{Real + Synthetic} & - & \cellcolor{gray!20}\textbf{9.98} & \cellcolor{gray!20}\textbf{0.80} & \cellcolor{gray!20}\textbf{10.03} & \cellcolor{gray!20}\textbf{0.80} & \cellcolor{gray!20}\textbf{3.30} & \cellcolor{gray!20}\textbf{0.93} & \cellcolor{gray!20}\textbf{3.33} & \cellcolor{gray!20}\textbf{0.93} \\
            \bottomrule
        \end{tabular}
    }
\end{table}

\subsection{Ablation Study}
We conduct ablation studies to validate the effectiveness of our design choices, focusing on the freezing strategy of the VGGT backbone and the contribution of synthetic training data. We evaluate different experiment setting on HO3D dataset, and all results are reported in \autoref{tab:ablation_combined}.
\paragraph{Impact of Backbone Freezing.} 
We first investigate how different freezing strategies affect the model performance. 
As shown in the top section of \autoref{tab:ablation_combined}, fixing the backbone parameters leads to suboptimal performance. 
Specifically, the freezing none setting significantly outperforms the freeze all (24 layers) setting, reducing MPJPE by a large margin. 
This suggests that although the pre-trained VGGT features are robust, the domain gap between the pre-training task and 3D hand reconstruction is non-negligible. 
A full trainable VGGT backbone allow the model to adapt its feature representation to capture the geometric details required for accurate hand mesh recovery.
\paragraph{Effectiveness of Synthetic Data.} 
We further analyze the impact of incorporating our rendered synthetic dataset into the training pipeline. 
Comparing the \textit{Real Only} baseline with our final \textit{Real + Synthetic} setting in \autoref{tab:ablation_combined}, we observe a consistent performance boost across all metrics (e.g., MPVPE decreases significantly). 
We attribute this improvement to two factors: 
(1) The synthetic data provides perfect ground-truth annotations, mitigating the noise inherent in real-world labels; 
(2) The diverse poses and viewpoints in the synthetic set act as strong regularization, preventing the model from overfitting to the limited distribution of real-world data.

\begin{figure}[h]
    \centering
    \vspace{-1em} 
    
    \begin{subfigure}[b]{0.9\linewidth}
        \centering
        \includegraphics[width=\linewidth]{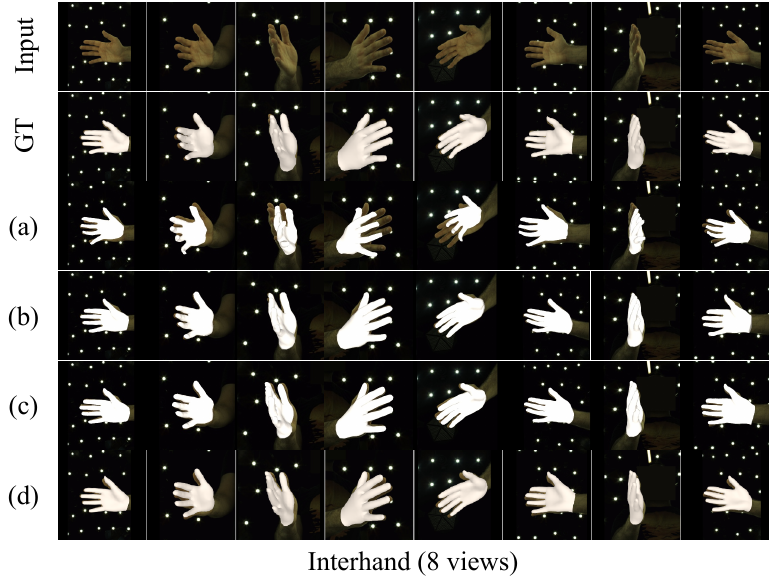}
        \label{fig:interhand}
    \end{subfigure}
    

    \begin{subfigure}[b]{0.9\linewidth}
        \centering
        \includegraphics[width=\linewidth]{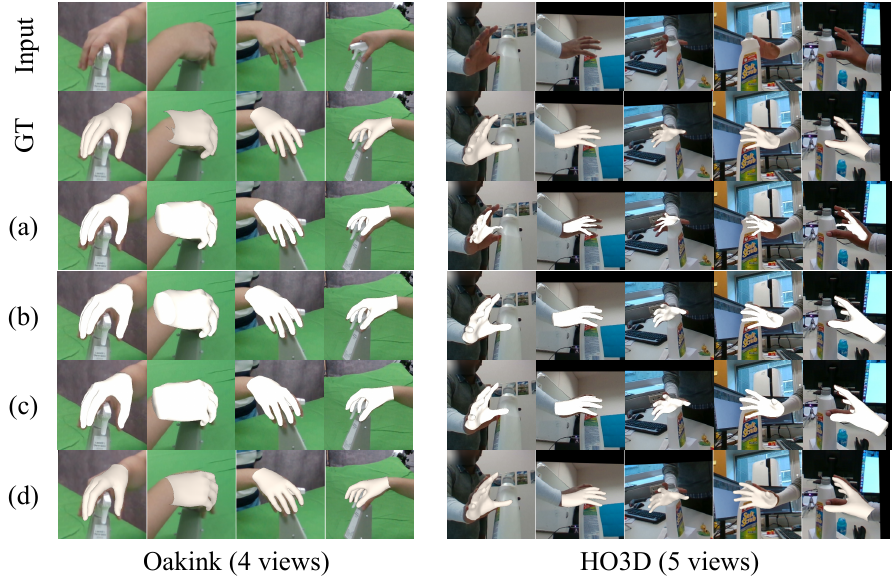}
        \label{fig:oakink_ho3d}
    \end{subfigure}
    \caption{\textbf{Qualitative comparison on InterHand2.6M, OakInk and HO3D.} We compare our method against baselines on open benchmarks. Rows correspond to: \textbf{Input} RGB, Ground Truth, \textbf{(a)}  Cameras Predicted by VGGT + \POEMg-large, \textbf{(b)} \POEMg-large, \textbf{(c)} Cameras Predicted by Ours  + \POEMg-large, and \textbf{(d)} Ours (Full).}
    \label{fig:qualitative_comparison}
    \vspace{-5pt}
\end{figure}

\subsection{Qualitative Comparison}
\autoref{fig:qualitative_comparison} visualizes the reconstruction results of our method compared to existing baselines on the InterHand2.6M~\cite{Moon_2020_ECCV_InterHand26M}, OakInk~\cite{yang2022oakink}, and HO3D~\cite{hampali2020honnotate,hampali2022keypointtransformer} datasets. For more comprehensive visualizations across additional datasets, please refer to the Supplementary Material.
As shown in~\autoref{fig:qualitative_comparison}, the baseline using VGGT-predicted cameras (row a) often suffers from significant misalignment due to inaccurate camera estimation. 
In contrast, our full method (row d) achieves reconstruction quality comparable to the baseline using Ground Truth cameras (row b). 
This demonstrates that our approach effectively recovers accurate camera parameters and hand geometry, producing tight alignment and physically plausible poses without relying on ground truth intrinsics.

\begin{figure}[h]
    \centering
    \includegraphics[width=\linewidth]{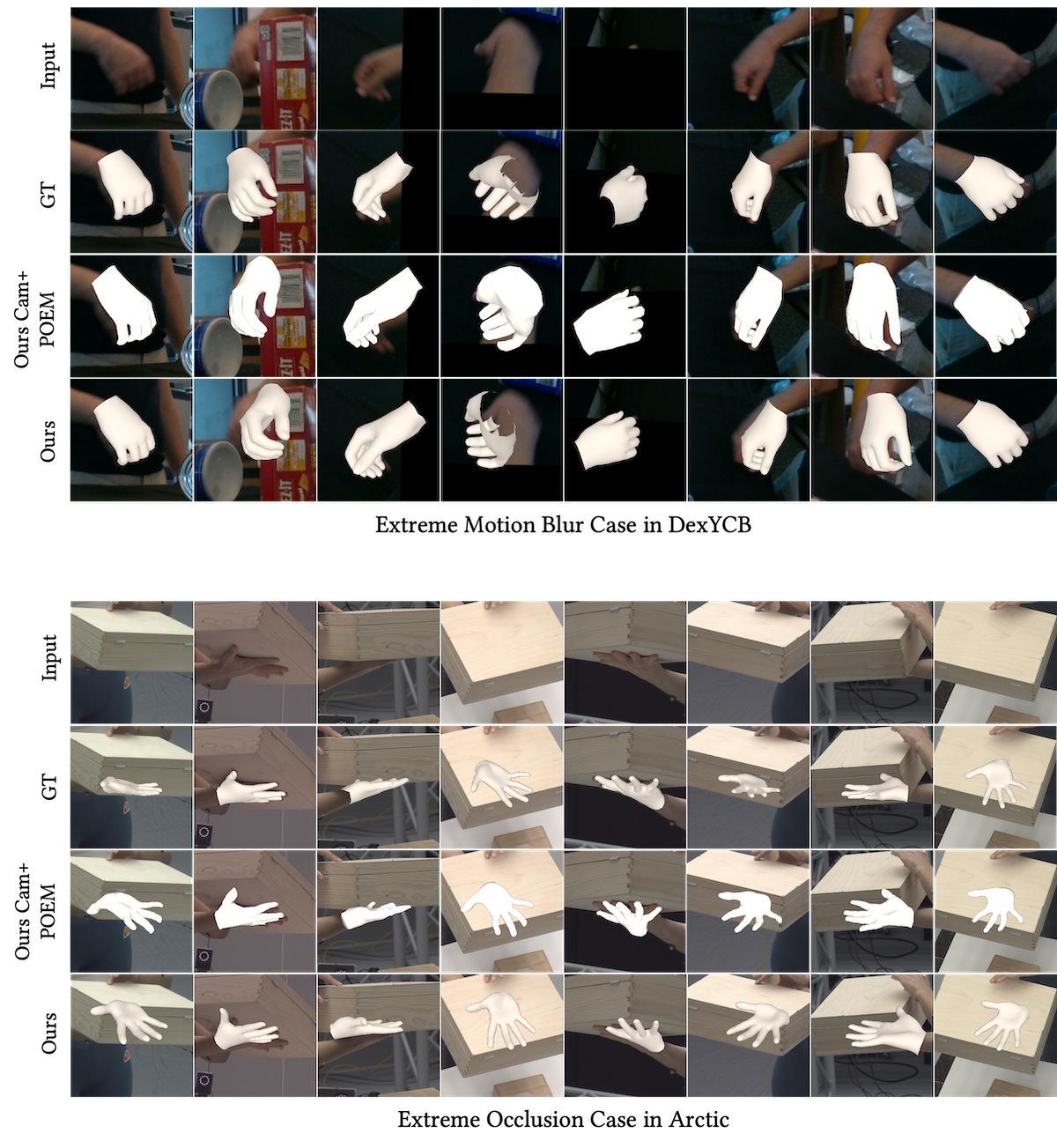}
    \vspace{-5pt}
    \caption{\textbf{Robustness in Extreme Scenarios.} Our method maintains reasonable reconstruction quality even under challenging conditions, such as severe motion blur (top) and heavy object occlusion where the right hand is visible in only 2 out of 8 views (bottom).}
    \vspace{-15pt}
    \label{fig:extreme_cases}
\end{figure}

To evaluate the robustness of our framework, we test it under extreme conditions as shown in~\autoref{fig:extreme_cases}. 
Despite the challenges posed by severe motion blur (top row), where high-frequency edge details are degraded by rapid hand movements, our method successfully recovers the global hand pose and maintains structural integrity.
Furthermore, our approach demonstrates remarkable stability under heavy occlusion (bottom row). 
In scenarios where the hand is heavily obstructed by large objects our model effectively leverages the sparse visual cues combined with learned priors.

\subsection{Limitation and Future Work}
Similar to state-of-the-art methods like HaMeR~\cite{pavlakos2024reconstructing} and POEM~\cite{yang2024POEM}, our framework currently relies on off-the-shelf 2D detectors (e.g., ViTPose~\cite{xu2022vitpose}) for hand localization and cropping in unconstrained settings.
Furthermore, as the estimated camera parameters are defined within a normalized coordinate space, recovering the absolute metric depth of the wrist from uncalibrated images remains an open challenge. Addressing metric scale recovery from uncalibrated views represents a promising direction for future research.
Beyond isolated hand reconstruction, our large-scale synthetic data offers rich geometric annotations that are often scarce in real-world hand-object interaction datasets. We hope that both the proposed HGGT framework and this dataset can contribute meaningfully to future efforts in broader tasks, such as robotic grasping and human-object interaction reconstruction.

\section{Conclusion}
In this work, we introduced the first feed-forward framework for calibration-free multi-view hand reconstruction. Supported by our mixed-data training strategy and our newly created large-scale synthetic dataset of rendered multi-view hand-object data, our model demonstrates superior robustness and generalization, outperforming state-of-the-art methods on uncalibrated scenarios.

\clearpage

%
%
\bibliographystyle{splncs04}
\bibliography{main}

\clearpage
\setcounter{page}{1}

\appendix

\begin{center}
{\Large\textbf{HGGT: Robust and Flexible 3D Hand Mesh Reconstruction from Uncalibrated Images}}\\[0.5em]
{\large\textbf{Supplementary Material}}\\[1em]
\end{center}

\section{Implementation Details}
To facilitate future research and ensure the reproducibility of our work, we are committed to open-sourcing our complete pipeline. We will publicly release:
\begin{itemize}
    \item The complete training and evaluation code of our model.
    \item The large-scale multi-view synthetic dataset introduced in this work.
    \item The data generation pipeline used to synthesize the dataset.
\end{itemize}

\subsection{Network Architecture.} 
Our multi-view 3D hand reconstruction framework builds upon the Visual Geometry Grounded Transformer (VGGT) architecture. The backbone employs a DINOv2 ViT-L/14~\cite{oquab2023dinov2} encoder for patch embedding, followed by a VGGT aggregator with \(24\) transformer blocks. The aggregator operates with an embedding dimension of \(1024\), \(16\) attention heads, and an MLP expansion ratio of \(4.0\). It utilizes alternating attention between frame-level and global-level tokens with rotary position embeddings (RoPE frequency=\(100\)). Each view maintains \(4\) register tokens alongside camera and patch tokens.
For hand pose estimation, we introduce a unified cross-attention refinement module, which takes hand tokens, camera tokens and concatenated image features from the aggregator and processes them through \(4\) cross-attention blocks, each with \(16\) attention heads and an MLP ratio of \(4\). 

\subsection{Training Setup.} 
In addition to the optimization settings detailed in the main text, we employ mixed-precision training with BF16 automatic mixed precision (AMP) to improve memory efficiency. Furthermore, we apply gradient clipping with an L2 norm threshold of \(1.0\) across all modules to ensure training stability. 

\subsection{Loss Weights.} 
Our multi-task loss function combines several objectives with carefully tuned weights. For hand pose estimation, we use a hand loss weight of \(1.0\), with sub-weights of \(0.1\) for pose parameters, \(0.05\) for shape coefficients, \(5.0\) for 3D keypoints, \(1.0\) for single-view 2D keypoints, and \(10.0\) for multi-view 2D keypoint consistency. The camera loss receives a weight of \(5.0\). Crucially, we employ intermediate supervision across the \(4\) cross-attention blocks. The hand and camera losses are computed at the output of each block and aggregated using exponentially increasing stage weights to emphasize the later refinement stages. In contrast, the 2D projection loss is applied exclusively to the final block's output to ensure terminal geometric consistency without disrupting the intermediate feature learning.

\section{Additional Ablation Studies}

In this section, we provide further ablation studies to validate the specific design choices in our network architecture and training strategy. The quantitative results are summarized in~\autoref{tab:supp_ablation}.

\begin{table}[h]
\centering
\caption{Additional ablation studies on the intermediate supervision strategy, projection loss placement, and the number of cross-attention blocks. PA-MPVPE and PA-MPJPE are reported in millimeters (mm).}
\label{tab:supp_ablation}
\resizebox{\linewidth}{!}{
\begin{tabular}{l c c c c}
\toprule
\textbf{Configuration} & \textbf{PA-MPVPE} \(\downarrow\) & \textbf{PA-MPJPE} \(\downarrow\) & \textbf{PA-AUC$_{V50}$} \(\uparrow\) & \textbf{PA-AUC$_{J50}$} \(\uparrow\) \\
\midrule
\multicolumn{5}{l}{\textit{Intermediate Supervision Strategy}} \\
\midrule
No Intermediate Sup. & 3.52 &3.53 &0.92 & 0.92\\
Constant Weights &  3.38& 3.37& 0.93& 0.92 \\
\rowcolor{gray!10} Exponential (Ours) &  3.30 & 3.33 & 0.93 & 0.93 \\
\midrule
\multicolumn{5}{l}{\textit{Number of Cross-Attention Blocks}} \\
\midrule
2 Blocks &  4.51& 4.55& 0.87& 0.86 \\
3 Blocks &  3.67& 3.70& 0.90& 0.90\\
\rowcolor{gray!10} 4 Blocks (Ours) &  3.30 & 3.33 & 0.93 & 0.93 \\
\midrule
\multicolumn{5}{l}{\textit{Placement of 2D Projection Loss}} \\
\midrule
All Blocks &  3.47 &3.45 &0.91 &0.91   \\
\rowcolor{gray!10} Final Block Only (Ours) &  3.30 & 3.33 & 0.93 & 0.93  \\
\bottomrule
\end{tabular}
}
\end{table}

\subsection{Intermediate Supervision Strategy}
As described in the main text, we apply intermediate supervision across the \(4\) cross-attention blocks using exponentially increasing stage weights. To validate this design, we compare our approach with two baselines: (1) \textbf{No Intermediate Supervision}, where losses are only applied to the final output, and (2) \textbf{Constant Weights}, where all intermediate blocks receive equal loss weights (\textit{i.e.}, \(1.0\)). As shown in~\autoref{tab:supp_ablation}, the exponentially increasing strategy yields the best performance. Constant weights tend to over-constrain the early layers, while our strategy effectively guides the progressive refinement without disrupting early feature extraction.

\subsection{Number of Cross-Attention Blocks}
We also investigate the impact of the number of cross-attention blocks in our refinement module. We test configurations with \(2\), \(3\), and \(4\) blocks. As reported in ~\autoref{tab:supp_ablation}, increasing the number of blocks  brings improvements in hand reconstruction accuracy, demonstrating the scalability of our method.

\subsection{Placement of 2D Projection Loss}
A key detail in our training setup is applying the 2D projection loss exclusively to the final block's output. We ablate this by comparing it against a variant where the 2D projection loss is applied to all intermediate blocks. The results in~\autoref{tab:supp_ablation} demonstrate that applying the projection loss to all layers degrades the performance. This confirms our hypothesis that enforcing strict 2D geometric consistency in early layers disrupts the intermediate 3D feature learning, whereas applying it only at the terminal stage ensures final alignment without feature distortion.

\section{More Qualitative Visualizations }
We provide additional qualitative results in this section to further demonstrate the effectiveness and robustness of our proposed method. These extensive visualizations showcase our model's capability in accurately predicting camera poses and reconstructing complex hand-object interactions across diverse and challenging scenarios.

\subsection{Open Benchmark.}

\begin{figure}[h]
\centering

\begin{minipage}{\linewidth}
    \centering
    \includegraphics[width=0.9\linewidth]{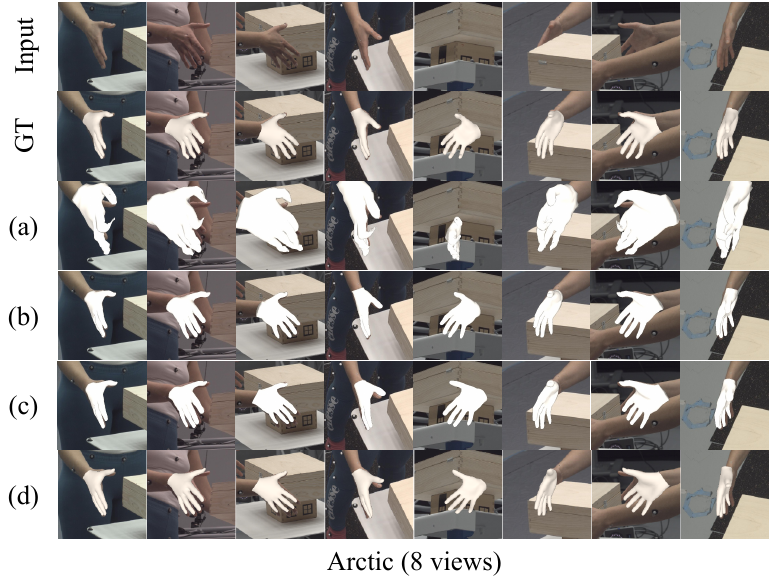}
        \end{minipage}

\begin{minipage}{\linewidth}
    \centering
    \includegraphics[width=0.9\linewidth]{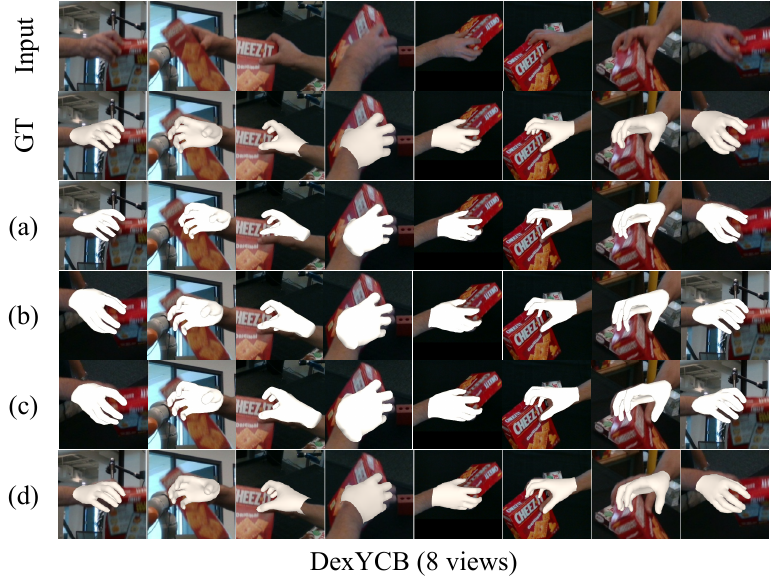}
\end{minipage}

\caption{\textbf{Qualitative comparison on the DexYCB dataset.} We compare our method against baselines on DexYCB dataset. Rows correspond to: Input, Ground Truth, \textbf{(a)} Cameras Predicted by VGGT + \POEMg-large, \textbf{(b)} \POEMg-large, \textbf{(c)} Cameras Predicted by Ours + \POEMg-large, and \textbf{(d)} Ours (Full).}
\label{fig:supp_comp_arctic_dexycb}

\end{figure}

In this subsection, we present comprehensive visual comparisons on several widely used open benchmarks. Specifically,~\autoref{fig:supp_comp_arctic_dexycb} illustrates the qualitative results on the ARCTIC and DexYCB datasets, respectively. In these challenging scenarios, our full pipeline consistently yields better alignment and more accurate 3D reconstructions compared to the baselines. 

\begin{figure}[h]
    \centering
    \includegraphics[width=\linewidth]{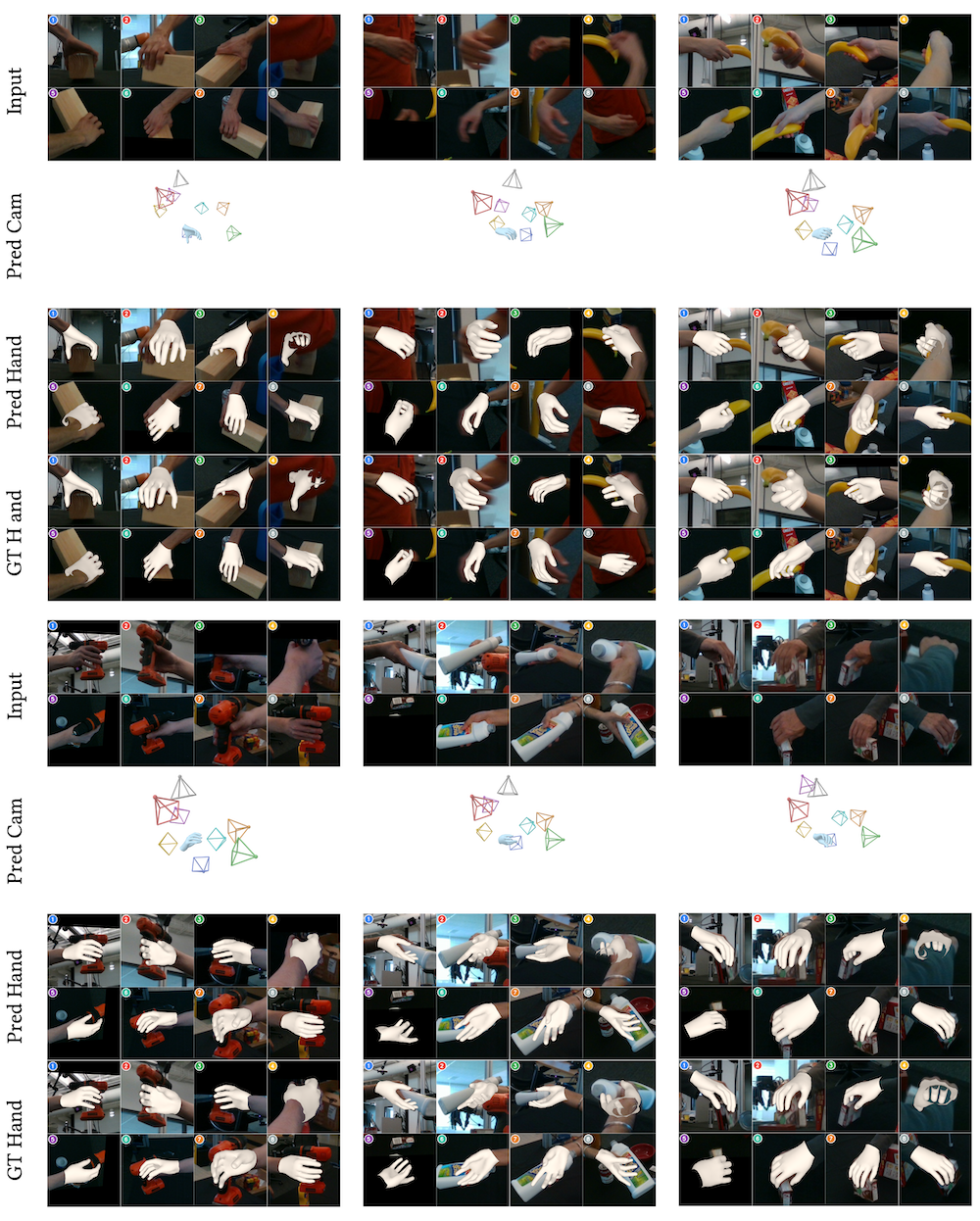}
    \caption{Additional qualitative results on the DexYCB~\cite{chao2021dexycb} dataset.}
    \label{fig:supp_dexycb_more}
\end{figure}

\begin{figure}[h]
    \centering
    \includegraphics[width=\linewidth]{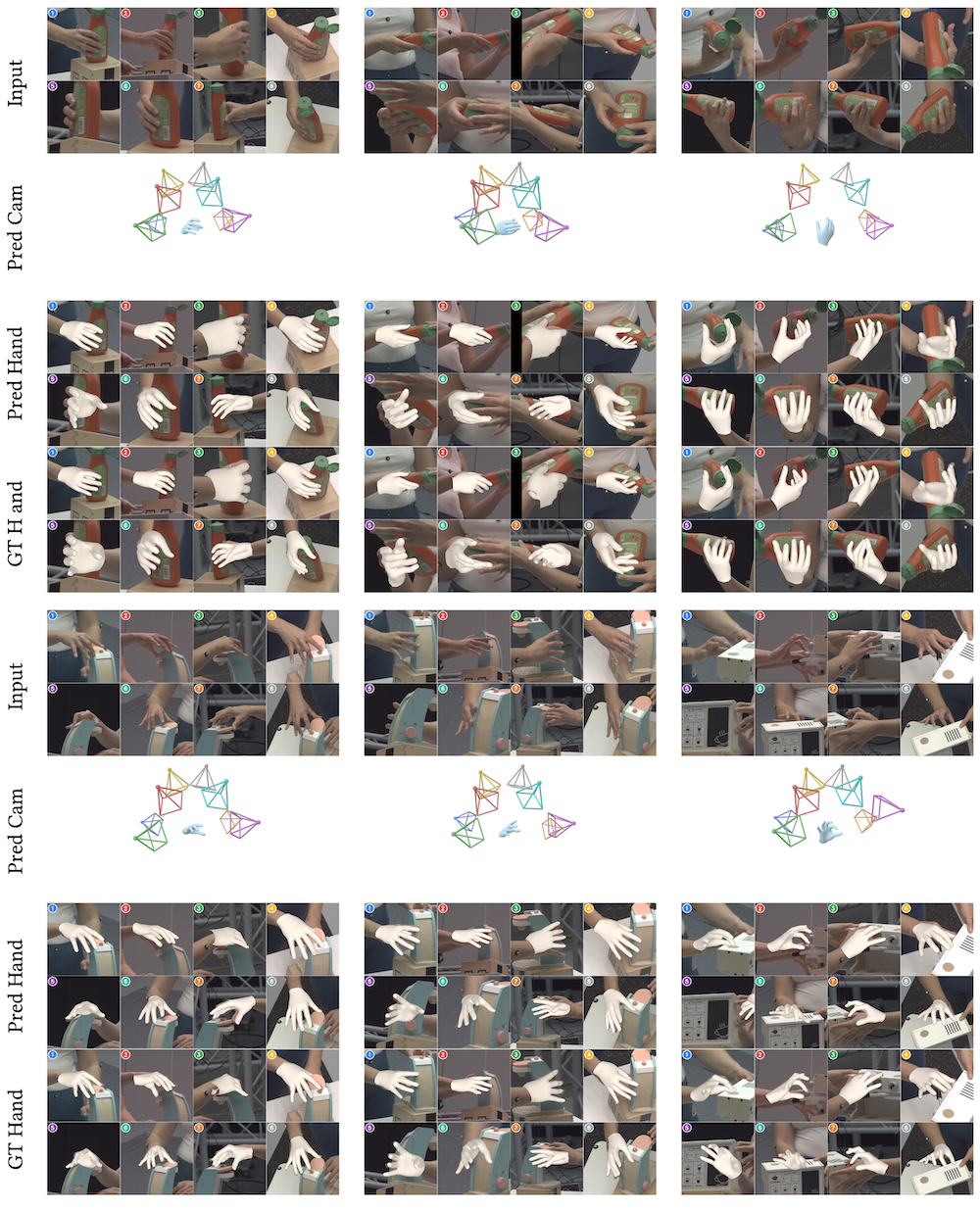}
    \caption{Additional qualitative results on the Arctic~\cite{fan2023arctic} dataset.}
    \label{fig:supp_arctic_more}
\end{figure}

\begin{figure}[h]
    \centering
    \includegraphics[width=\linewidth]{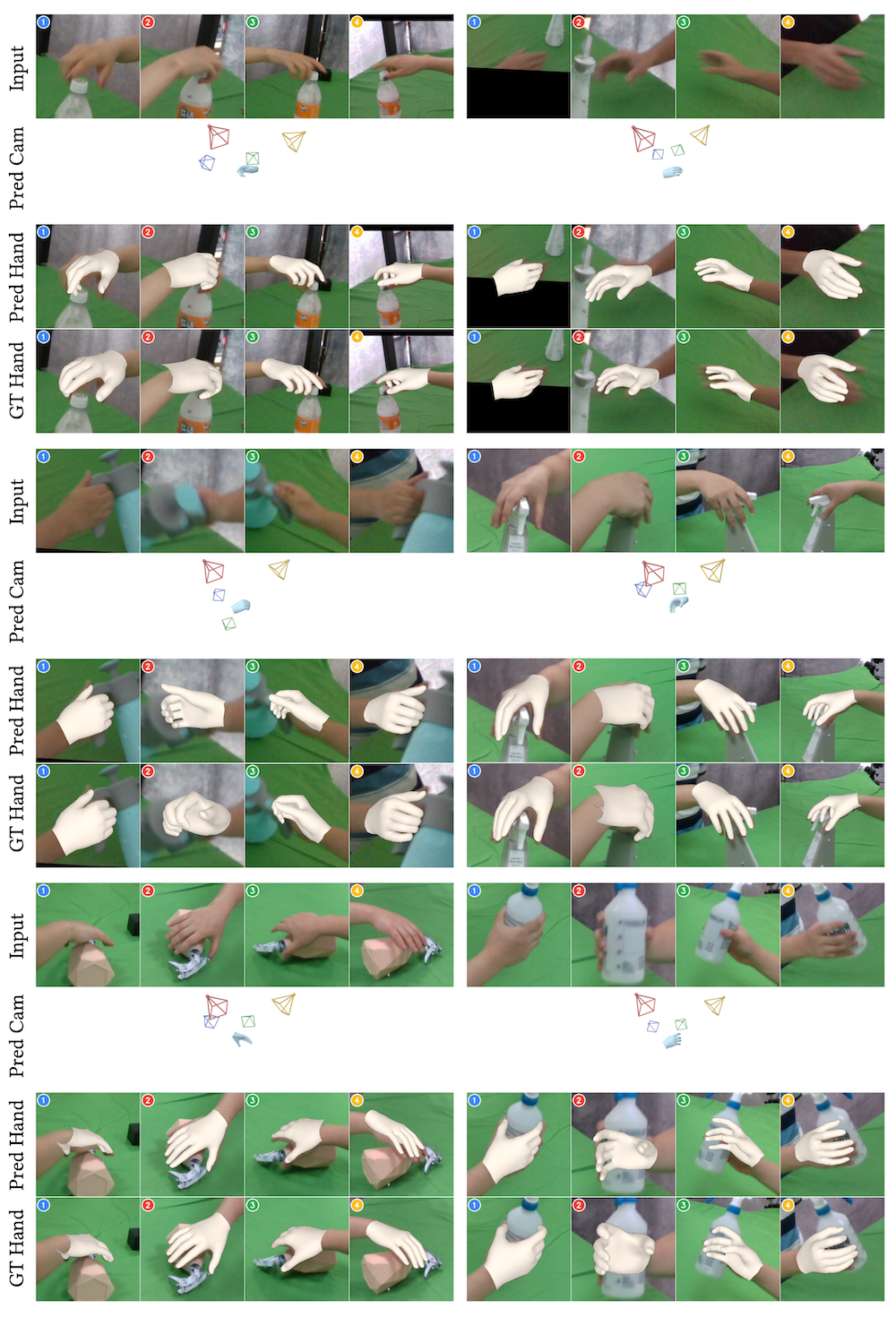}
    \caption{Additional qualitative results on the OakInk~\cite{yang2022oakink} dataset.}
    \label{fig:supp_oakink_more}
\end{figure}

\begin{figure}[h]
    \centering
    \includegraphics[width=\linewidth]{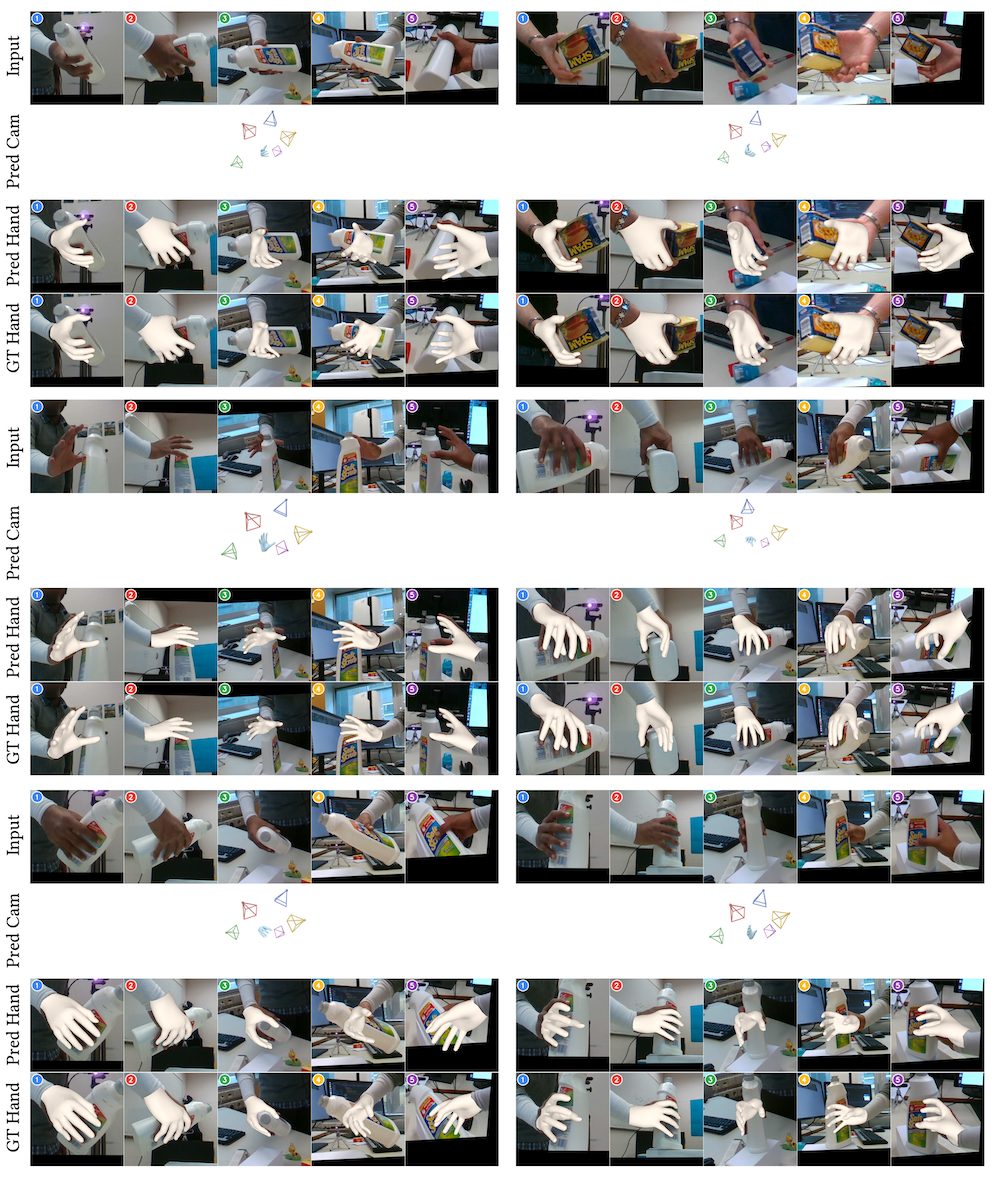}
    \caption{Additional qualitative results on the HO3D~\cite{hampali2020honnotate} dataset.}
    \label{fig:supp_ho3d_more}
\end{figure}

\begin{figure}[h]
    \centering
    \includegraphics[width=\linewidth]{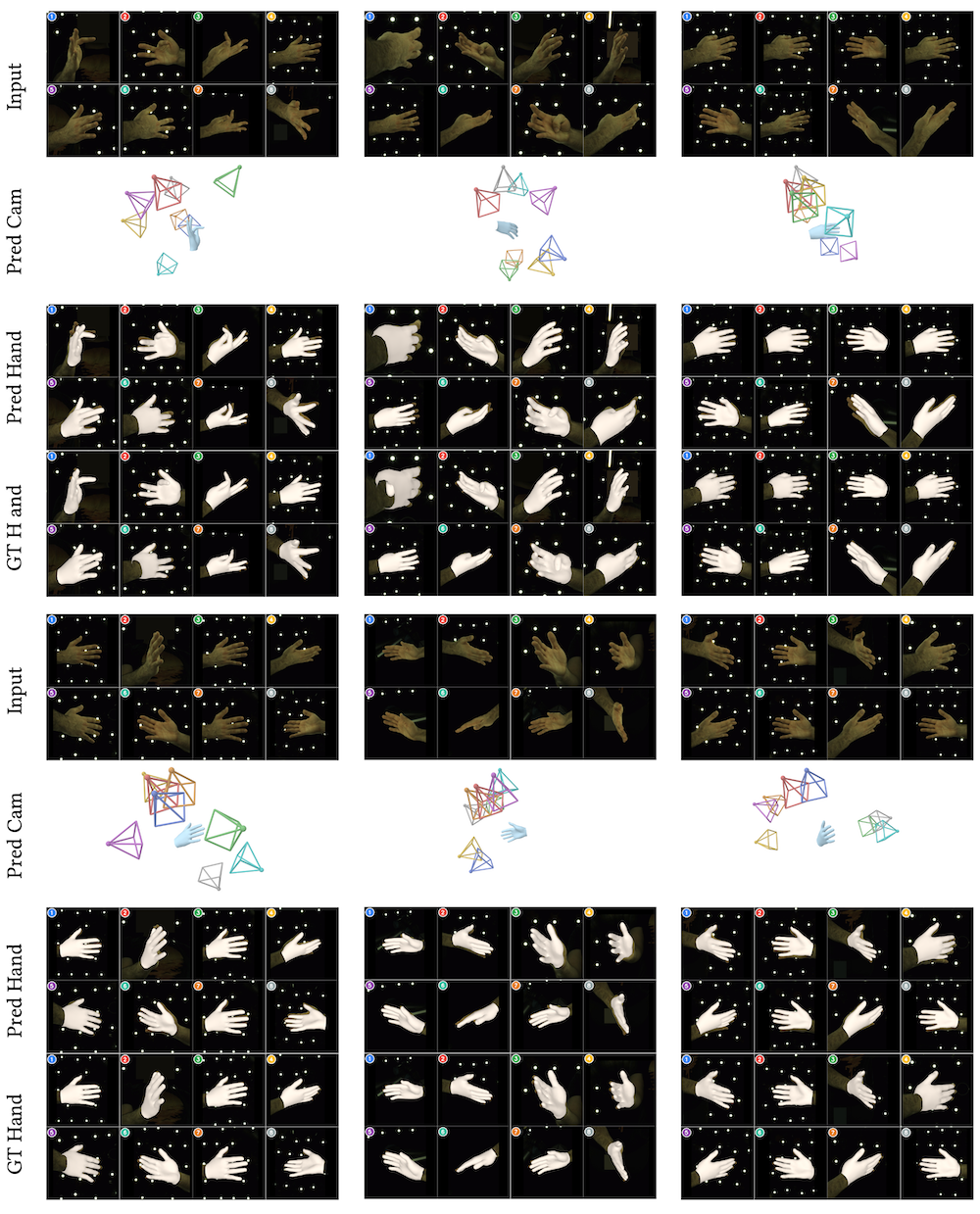}
    \caption{Additional qualitative results on the InterHand2.6M~\cite{Moon_2020_ECCV_InterHand26M} dataset.}
    \label{fig:supp_interhand_more}
\end{figure}

Furthermore,~\autoref{fig:supp_oakink_more}, ~\autoref{fig:supp_dexycb_more}, ~\autoref{fig:supp_arctic_more}, ~\autoref{fig:supp_ho3d_more} and ~\autoref{fig:supp_interhand_more} provide additional qualitative visualizations on open dataset. These extra examples further demonstrate our model's robustness when handling diverse scenarios.

\subsection{In-the-wild Video Inference}
\begin{figure}[htbp]
    \centering
    \includegraphics[width=\linewidth]{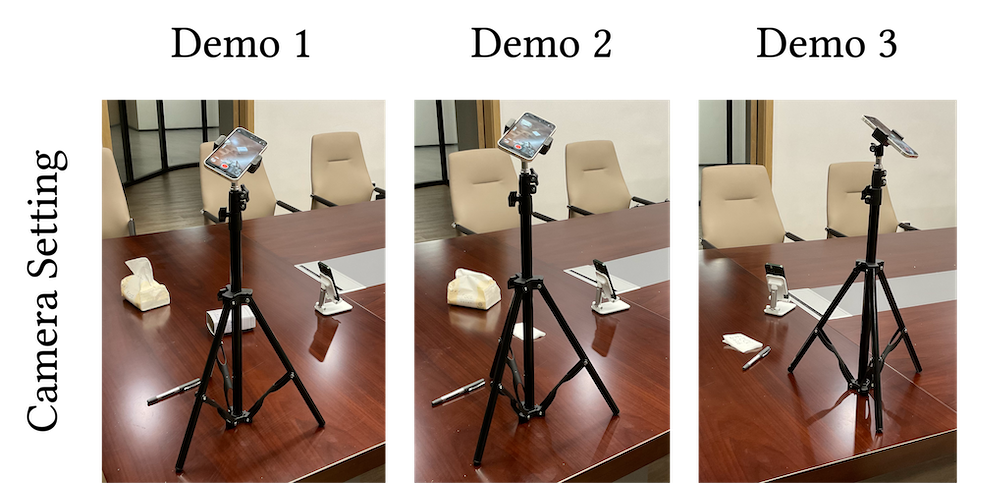} 
    \caption{Camera settings for in-the-wild video capture. The videos are recorded using two smartphones placed at random positions without prior calibration.}
    \label{fig:supp:camera_settings}
\end{figure}

To further demonstrate the practical applicability of our method in real-world scenarios, we evaluate it on continuous in-the-wild video sequences. As shown in \autoref{fig:supp:camera_settings}, the videos are captured using a casual setup consisting of only two smartphones placed at random positions. 

\begin{figure}[htbp]
    \centering
    \includegraphics[width=\linewidth]{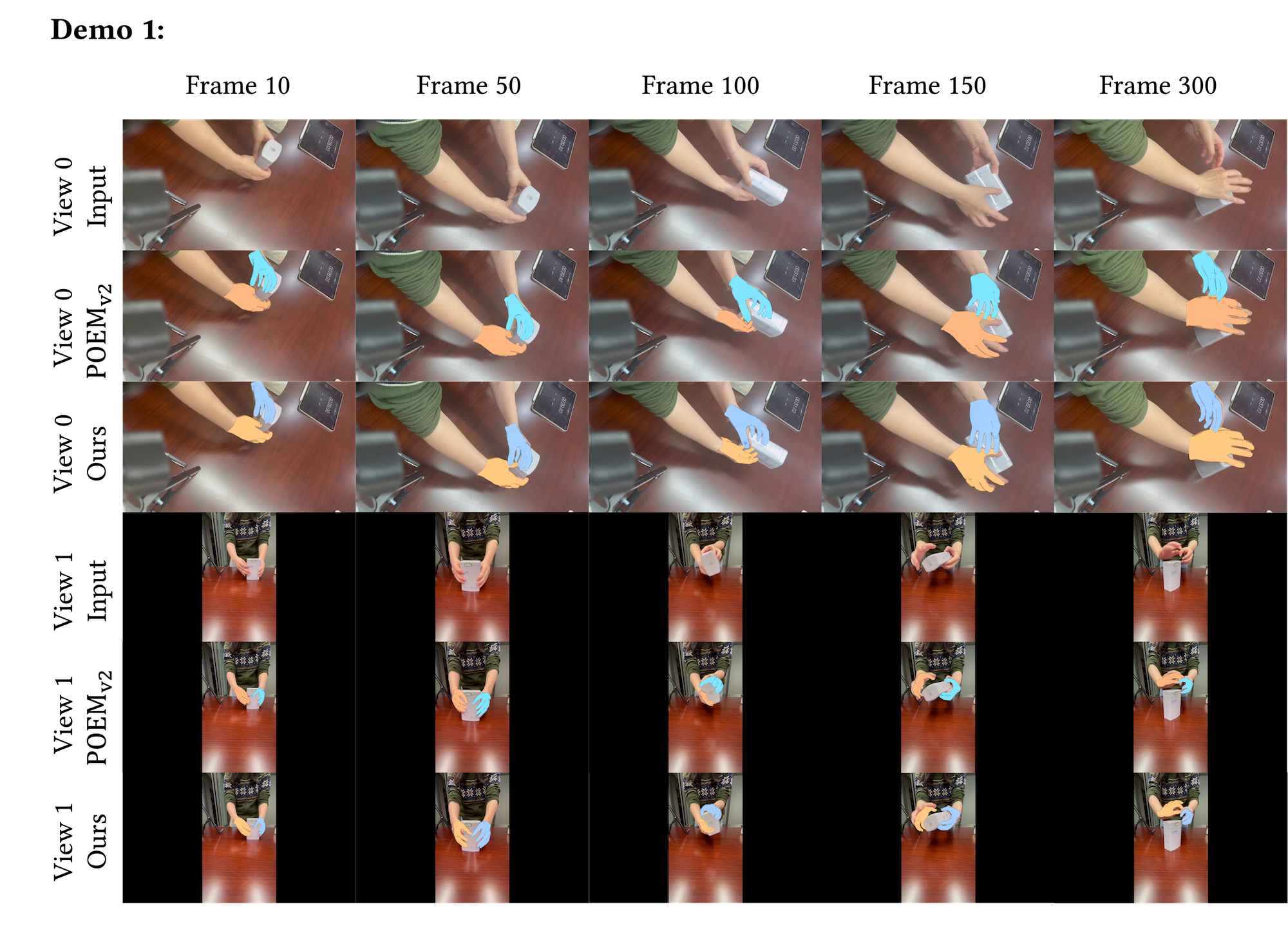} 
    \caption{Qualitative comparison with \POEMg-large on in-the-wild video sequence (Demo 1).}
    \label{fig:supp:real_demo1}
\end{figure}

\begin{figure}[htbp]
    \centering
    \includegraphics[width=\linewidth]{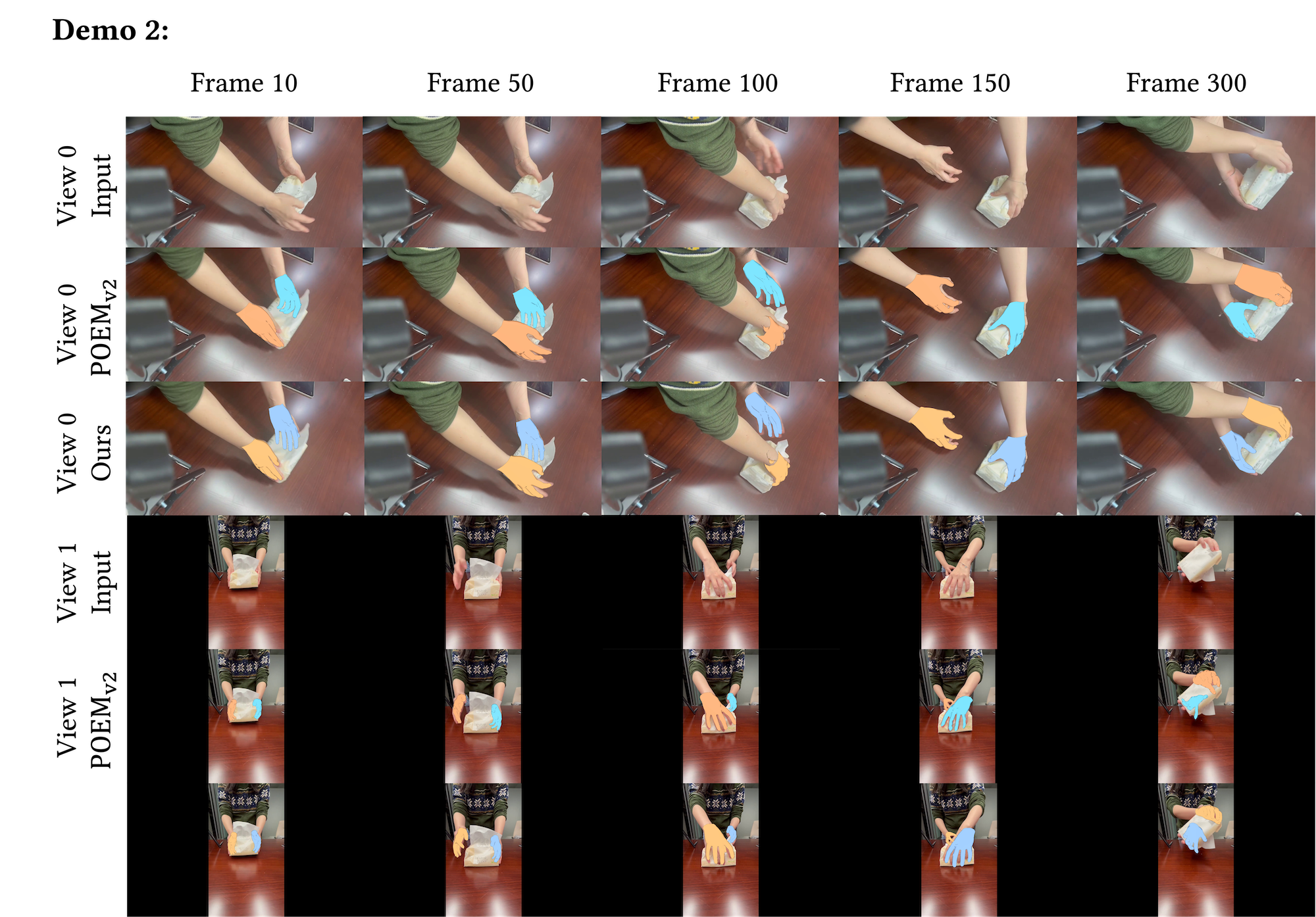} 
    \caption{Qualitative comparison with \POEMg-large on in-the-wild video sequence (Demo 2).}
    \label{fig:supp:real_demo2}
\end{figure}

\begin{figure}[htbp]
    \centering
    \includegraphics[width=\linewidth]{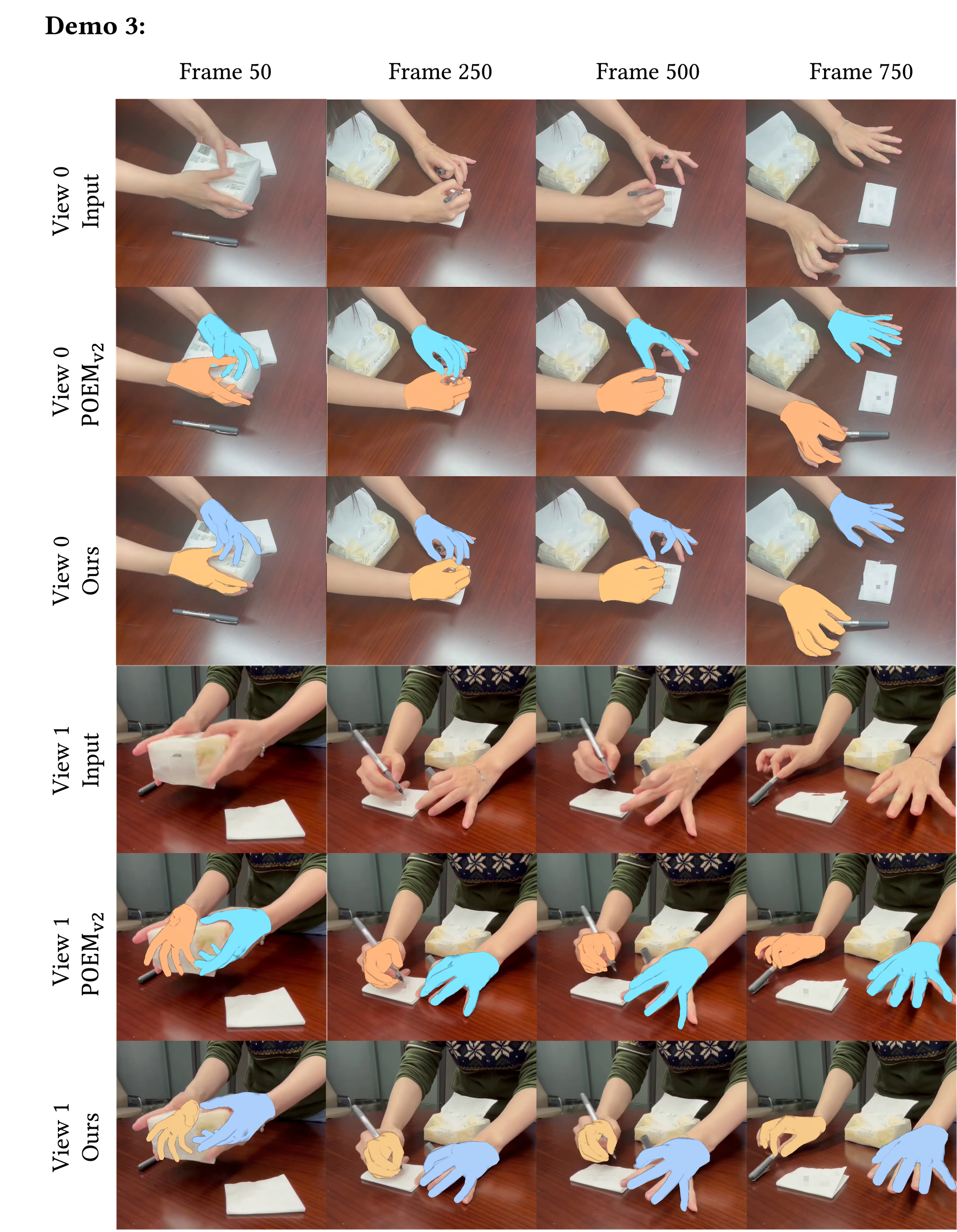} 
    \caption{Qualitative comparison with \POEMg-large on in-the-wild video sequence (Demo 3).}
    \label{fig:supp:real_demo3}
\end{figure}

Given the input videos, our model processes them frame-by-frame to predict the 3D hand reconstructions. To highlight the robustness of our approach, we compare our results with \POEMg-large. Since \POEMg-large strictly requires calibrated cameras, we supply it with the camera parameters predicted by our method to enable the comparison. The qualitative comparisons across three different demo sequences are illustrated in \autoref{fig:supp:real_demo1}, \autoref{fig:supp:real_demo2}, and \autoref{fig:supp:real_demo3}. Please refer to the  video in our project page for dynamic visualization results. 

Notably, the inference of our HGGT is highly efficient, requiring only 0.23 seconds per frame. However, our current pipeline relies on an external hand detector to obtain initial bounding boxes, which introduces an additional processing time of approximately 1.2 seconds per frame. We acknowledge this dependency as a limitation of the current system and plan to address it in future work by integrating a more efficient, end-to-end detection module to further accelerate the overall pipeline.

\end{document}